\documentclass{article}

\usepackage[preprint]{neurips_2026}

\usepackage[utf8]{inputenc}
\usepackage[T1]{fontenc}
\usepackage{hyperref}
\usepackage{url}
\usepackage{booktabs}
\usepackage{amsmath}
\usepackage{amsfonts}
\usepackage{nicefrac}
\usepackage{microtype}
\usepackage[table]{xcolor}
\usepackage{multirow}
\usepackage{graphicx}
\usepackage{fvextra}          %
\usepackage{algorithm}
\usepackage{algorithmic}
\usepackage{xspace}
\usepackage{enumitem}

\newcommand{\methodfull}{Agentic Conversation Context Curation\xspace}
\newcommand{\method}{AC3\xspace}

\newcommand{\logogpt}{\raisebox{-0.20ex}{\includegraphics[height=1.7ex]{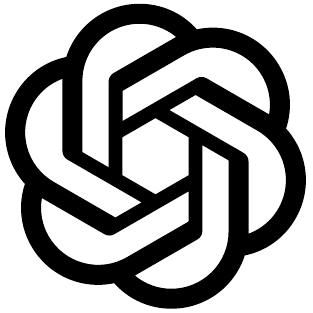}}}
\newcommand{\logodsv}{\raisebox{-0.20ex}{\includegraphics[height=1.7ex]{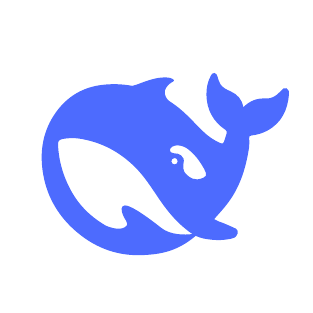}}}
\newcommand{\logokimi}{\raisebox{-0.20ex}{\includegraphics[height=1.7ex]{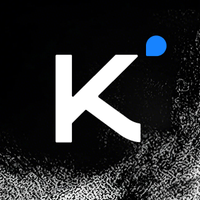}}}
\newcommand{\mgpt}{\logogpt\,gpt-5.4}
\newcommand{\mdsv}{\logodsv\,dsv4f}
\newcommand{\mkimi}{\logokimi\,k2.6}

\newcounter{promptctr}
\newcommand{\promptinput}[2][]{%
  \VerbatimInput[
    fontsize=\footnotesize,
    breaklines=true,
    breakanywhere=true,
    breaksymbolright=\small$\hookrightarrow$,
    obeytabs=true,
    tabsize=2,
    frame=leftline,
    framerule=0.6pt,
    rulecolor=\color{gray},
    #1
  ]{#2}%
}

\definecolor{darkblue}{rgb}{0, 0, 0.5}
\hypersetup{colorlinks=true, citecolor=darkblue, linkcolor=darkblue, urlcolor=darkblue}

\title{Agentic Context Management for Multi-Turn Human--AI Conversations}

\author{%
  Matthew Ho$^{1}$\quad
  Chenglong Wang$^{2}$\quad
  Baolin Peng$^{2}$\quad
  Nebojsa Jojic$^{2}$\quad
  Philippe Laban$^{2}$\\
  \textbf{Michel Galley}$^{2}$\quad
  \textbf{Jianfeng Gao}$^{2}$\quad
  \textbf{Lianhui Qin}$^{1}$\\[4pt]
  $^{1}$University of California, San Diego\qquad $^{2}$Microsoft Research\\
  \small\texttt{\{mah048,lianhui\}@ucsd.edu}\\
  \small\texttt{\{chenglong.wang,baolinpeng,jojic,plaban,mgalley,jfgao\}@microsoft.com}
}

\definecolor{tshade0}{HTML}{E0F1F4}
\definecolor{tshade1}{HTML}{C6E5EB}
\definecolor{tshade2}{HTML}{ABD9E1}
\definecolor{tshade3}{HTML}{91CDD8}
\definecolor{tshade4}{HTML}{77C2CF}
\definecolor{tshade5}{HTML}{5CB6C6}
\definecolor{tshade6}{HTML}{42AABC}

\newcommand{\score}[1]{%
\ifnum#1<20 \cellcolor{tshade0}\else%
\ifnum#1<35 \cellcolor{tshade1}\else%
\ifnum#1<50 \cellcolor{tshade2}\else%
\ifnum#1<65 \cellcolor{tshade3}\else%
\ifnum#1<78 \cellcolor{tshade4}\else%
\ifnum#1<88 \cellcolor{tshade5}\else%
\cellcolor{tshade6}\fi\fi\fi\fi\fi\fi%
#1}
\newcommand{\scored}[2]{%
\ifnum#1<20 \cellcolor{tshade0}\else%
\ifnum#1<35 \cellcolor{tshade1}\else%
\ifnum#1<50 \cellcolor{tshade2}\else%
\ifnum#1<65 \cellcolor{tshade3}\else%
\ifnum#1<78 \cellcolor{tshade4}\else%
\ifnum#1<88 \cellcolor{tshade5}\else%
\cellcolor{tshade6}\fi\fi\fi\fi\fi\fi%
#2}

\begin{document}

\maketitle

\begin{abstract}
Large language models degrade in multi-turn conversation partly because of \emph{context pollution}: the model's own earlier outputs stay in context and anchor later turns, so an early misreading of the user's intent can persist after the user corrects it. Assistant omission (AO), which drops every assistant message, removes this pollution but fails when interaction is \emph{referential}: drafts, decisions, and tool-call results may exist only in assistant turns, and later turns refer back to them. We propose \textbf{\methodfull{} (\method)}, a training-free, inference-time method in which a separate analyzer (1)~consolidates the task specification from the user's messages alone, (2)~audits the assistant's prior outputs against it, and (3)~edits the context, either appending the analysis or discarding the assistant side and rebuilding the context from the specification. Building the analyzer exposed a second finding: pollution is \emph{contagious}. An analyzer that sees the assistant's reasoning while extracting the specification can do worse than no intervention at all, and instructing it to ignore them did not close the gap, so we exclude assistant turns structurally. On Lost in Conversation (replaying difficulty-selected instances), where AO is an oracle by construction, \method closes 55--84\% of the gap between full context and AO on three of four tasks and exceeds AO on the fourth, and an LLM that faithfully summarizes the history does not reproduce the gain even with more compute. On real human-AI conversations from WildChat, a GPT-5-mini judge prefers \method's next turn over both full context and AO in every configuration of our main WildChat table, although this preference depends on the judge; on CollabLLM the results are mixed. In stateful tool use (tau2-bench), AO collapses to 0\% on every model; \method does not collapse, but no \method operator beats full context. Detector studies show that Reset removes injected pollution but is not span-selective, which locates the gain in rebuilding the specification rather than in preserving selected assistant spans.
\end{abstract}

\section{Introduction}
\label{sec:intro}

\begin{figure}[t]
\centering
\includegraphics[width=\textwidth]{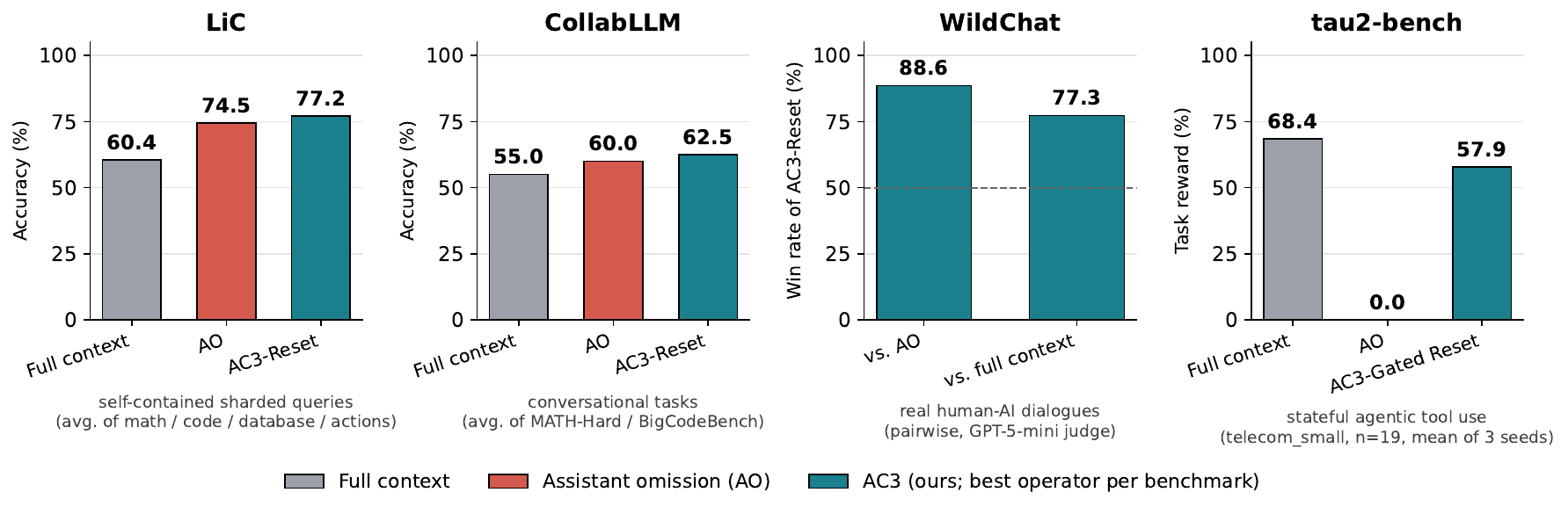}
\caption{\textbf{The value of omitting assistant turns falls as task referentiality grows.} Results for one respondent (gpt-5.4) on the four benchmarks, ordered by how much the next user turn depends on the assistant's prior turns. Values are taken from Table~\ref{tab:megatable} (LiC, CollabLLM, tau2-bench) and Table~\ref{tab:wildchat} (WildChat); the \method bar is the best operator for this respondent on each benchmark (on WildChat, best against AO). On LiC, where the user's messages together fully specify the task, AO is strong and \method is higher still. On CollabLLM, \method-Reset (62.5) is one problem out of 40 ahead of AO (60.0) and three ahead of full context (55.0). On WildChat, a GPT-5-mini judge prefers \method-Reset's next turn over both AO and full context (dashed line: parity). On tau2-bench, where tool-call results live only in assistant turns, AO collapses to 0\% and \method does not collapse, but full context remains best.}
\label{fig:story}
\end{figure}

Multi-turn interaction is the dominant mode of use for large language models, from coding assistants to computer-use agents such as OpenClaw~\citep{openclaw2026}. Yet LLMs perform substantially worse in multi-turn settings than on equivalent single-turn tasks: \citet{laban2025lost} report an average drop of 39\% across 15 models and 6 task domains, even in conversations of only $\sim$8 turns. They also identify the self-reinforcing dynamic behind the drop: early errors about the user's intent are recorded in the assistant's own messages and bias later turns, so errors compound instead of being corrected as new user information arrives. \citet{huang2026context} call this \emph{context pollution}, study it in real human-AI conversations, and show that a coarse fix works surprisingly well: \emph{assistant omission} (AO), which discards all assistant messages from the context.

AO works when the user's messages so far, taken together, contain enough information to complete the task; we call such turns \emph{self-contained}. Many turns are instead \emph{referential}: the correct response depends on content that exists only in earlier assistant turns. In plain chat, users build on what the assistant wrote (``sharpen sentence~2''; ``add a column to that query''). In tool use, the dependence is on environment state, a case we call \emph{stateful}: tool-call results and prior decisions are recorded only in assistant turns. For referential turns, removing assistant messages deletes the content that the next turn depends on.

\begin{figure}[t]
\centering
\includegraphics[width=\textwidth]{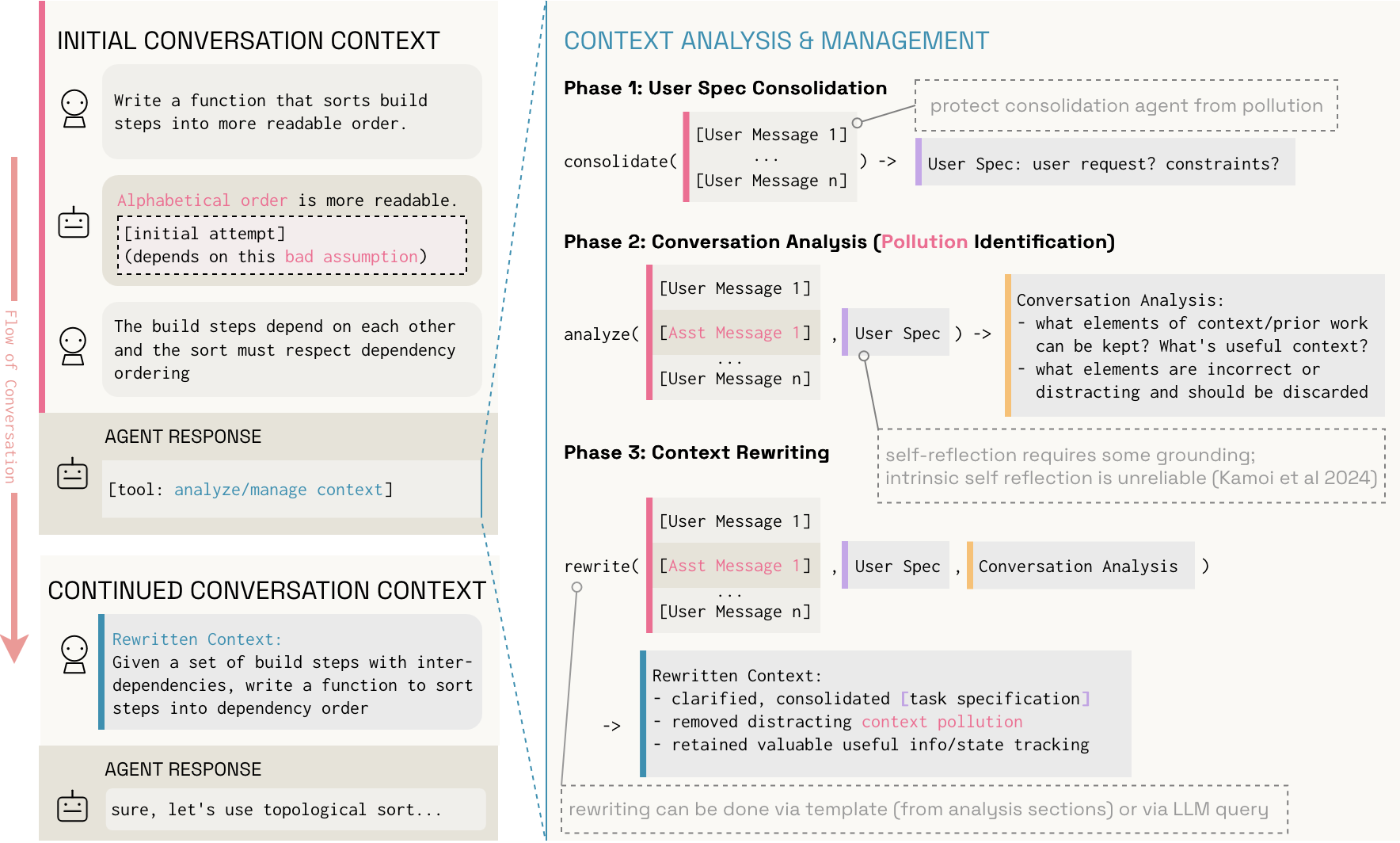}
\caption{\methodfull{} (\method). Before the assistant generates, a separate analyzer runs two analysis subtasks and one intervention step. Subtask~1 (specification extraction) sees only user turns, so it cannot be biased by the assistant's prior reasoning. Subtask~2 (approach evaluation) compares that specification against the assistant's full history. An operator (Augment, Reset, or Rewrite; the figure shows a rewritten context) then edits the context; the ``retained'' content in the rebuilt context is restated by the analyzer, not copied from assistant turns. Structural exclusion in Subtask~1 is essential: when the extractor sees assistant turns, accuracy falls below the full-context baseline on math and code (Table~\ref{tab:cognitive-hazard}).}
\label{fig:method}
\end{figure}

We propose \textbf{\methodfull{} (\method)}, a training-free, inference-time method that runs inside the agent harness before each assistant turn (Figure~\ref{fig:method}). A separate LLM, the \emph{analyzer}, (1)~extracts a consolidated task specification from the user's messages alone, (2)~audits the assistant's prior outputs against that specification, and (3)~edits the context with one of three operators: \emph{Augment} appends the analysis to the conversation, \emph{Reset} replaces the conversation with the specification and the prior work the analyzer judged correct, and \emph{Rewrite} has an LLM restructure the conversation around the specification. Augment, which only adds information, fires on every turn after a short warm-up; Reset and Rewrite, which remove content, act only when the analyzer flags an issue. Our default configuration, \emph{Gated Reset}, additionally skips the first user turns and caps the number of resets. With Reset, the mechanism is simple to state: \method detects pollution, discards the assistant side, and rebuilds the specification from the user side, so the assistant continues from what the user asked for rather than from its own invalidated reasoning. In stateful settings the analyzer also records the tool state it has observed, so that a rebuild can carry forward the state the next action needs.

Unlike pollution-oriented interventions such as AO and ERGO~\citep{khalid2025ergo}, which remove or rewrite context wholesale once triggered, \method audits the assistant's prior work against a user-grounded specification; unlike context management for long-horizon agents~\citep[][\emph{inter alia}]{schroeder2025thread,zhang2025rlm,sun2025contextfolding}, which compacts a trajectory to fit a token budget, it targets correctness in ongoing dialogue (Section~\ref{sec:related}). In spirit it is a use of test-time compute~\citep{wei2022chain,wang2023selfconsistency,yao2024tree,snell2024scalettc}, spent on curating the input context rather than on searching over outputs.

Building the analyzer exposed an unexpected finding: pollution is \emph{contagious}. A model tasked with auditing the conversation anchors on the assistant's prior reasoning if it can see it. When the specification extractor sees assistant turns, accuracy falls below that of no intervention on two of three LiC tasks, and asking the extractor to reason only from user-stated requirements does not close the gap (Section~\ref{sec:cognitive-hazard}; Appendix~\ref{app:soft-attention}). Context management must therefore apply to the context manager itself: each analyzer step sees only the conversation slices its role requires, and we enforce this structurally rather than by prompting.

We evaluate on four benchmarks ordered by referentiality (Figure~\ref{fig:story}). In Lost in Conversation (LiC)~\citep{laban2025lost}, each instance is a single-turn question sharded across turns, so the user's messages together fully specify the task and AO is a \emph{design oracle}. There, in replay mode on difficulty-selected instances, Gated Reset closes 55--84\% of the gap between full context and AO on math, code, and actions and exceeds AO on database (e.g., 15.8\%$\to$64.4\% on code, a mean of three reruns), and Reset improves over full context in 33 of 36 paired comparisons across three further respondent models on a second difficulty-selected pool. On real human-AI conversations from WildChat~\citep{zhao2024wildchat}, a GPT-5-mini judge prefers every \method configuration in Table~\ref{tab:wildchat} over both AO and full context, although the preference depends on the judge (Section~\ref{sec:discussion}). On CollabLLM~\citep{wu2025collabllm}, near the middle of the spectrum, results are mixed: relative to full context, Reset ranges from $-0.1$ to $+7.5$pp across respondents and AO from $0.0$ to $+5.0$pp, gaps of at most three problems out of 40. On stateful tool use in tau2-bench~\citep{barres2025tau2}, AO collapses to 0\% on every respondent because tool-call results live only in assistant turns. \method remains \emph{viable}, meaning it does not collapse (Gated Reset scores 7--12pp below full context, not significantly at $N{=}3$), but no \method operator beats full context there.

Our contributions are:
\begin{itemize}[leftmargin=*, topsep=2pt, itemsep=1pt, parsep=0pt]
    \item An account of when discarding assistant turns works (self-contained vs.\ referential turns) that explains AO's range from design oracle to 0\% (Section~\ref{sec:problem-setting}).
    \item \method, in which an analyzer shielded from the assistant's turns rebuilds the task specification from the user side and edits the context (Section~\ref{sec:methods}), and the finding that pollution is contagious, which makes that shielding a structural requirement (Section~\ref{sec:cognitive-hazard}).
    \item An evaluation on four benchmarks of increasing referentiality, with matched-compute baselines and a detector evaluation that locates the gain in the rebuild rather than in span-selective editing (Section~\ref{sec:results}).
\end{itemize}

\section{Problem setting}
\label{sec:problem-setting}

\begin{figure}[t]
\centering
\includegraphics[width=\textwidth]{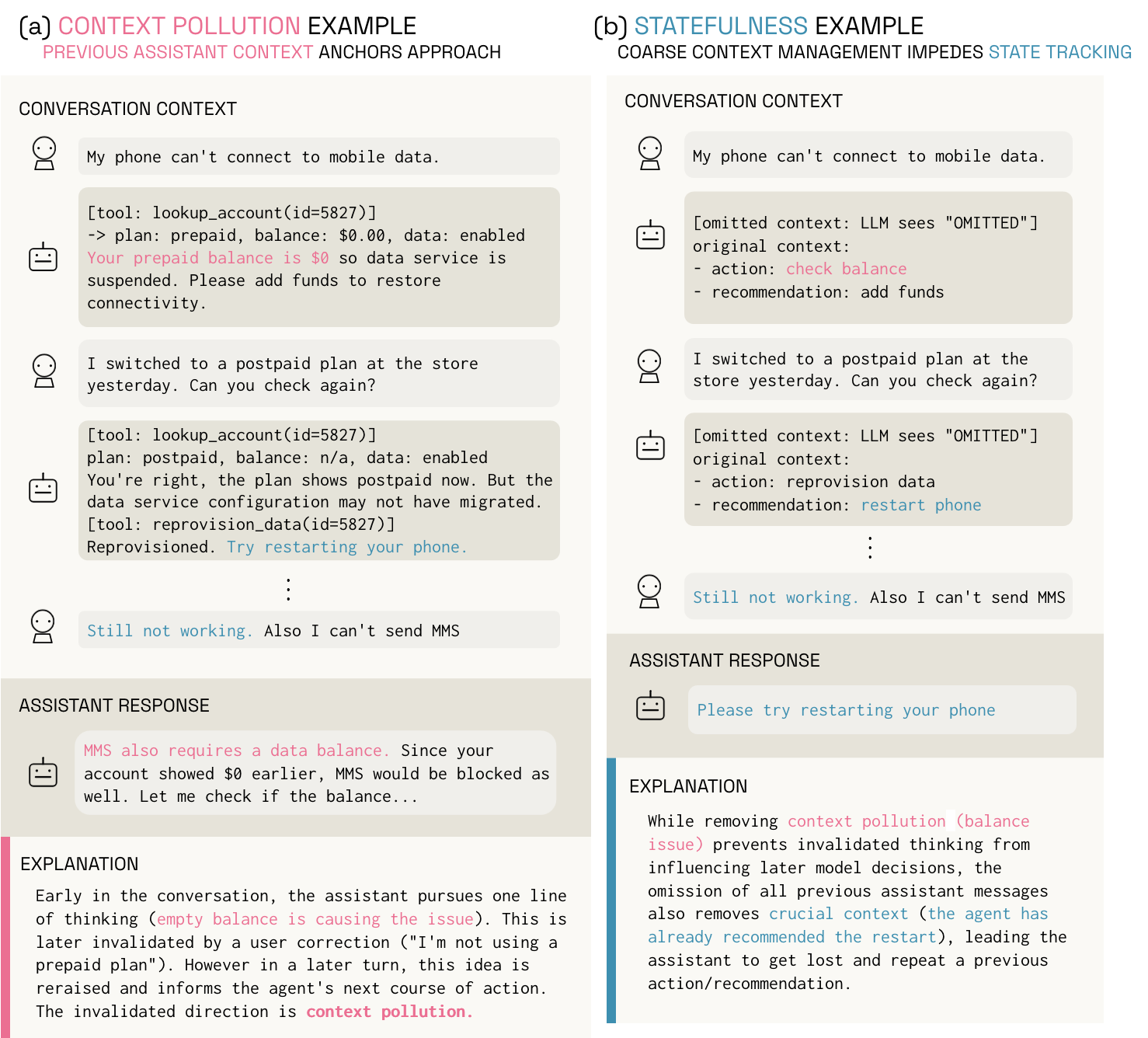}
\caption{Illustrative example. (a) Context pollution: an early incorrect interpretation by the assistant (a zero prepaid balance) stays in the context and re-anchors a later response even after the user has corrected it. (b) Coarse omission: AO removes the invalidated reasoning but also the state the next action needs (the restart the agent already recommended), so the agent repeats itself. The deficit is behavioral rather than a lack of capability: on LiC, the same models solve the task when it is posed in a single turn~\citep{laban2025lost}. Our method (Section~\ref{sec:methods}, Figure~\ref{fig:method}) supplies the missing control at the harness level by rebuilding the conversation around a specification drawn from the user's messages and dropping the invalidated reasoning.}
\label{fig:problem}
\end{figure}

\subsection{Setup and notation}
Consider a multi-turn interaction. At turn $t$, the conversation context is the message sequence $C_t = (u_1, a_1, u_2, a_2, \ldots, u_t)$, where $u_i$ and $a_i$ denote user and assistant messages respectively, and the assistant samples $a_t \sim p_\theta(\cdot \mid C_t)$. Let $U_t = (u_1, \ldots, u_t)$ denote the user prefix and $A_{t-1} = (a_1, \ldots, a_{t-1})$ the assistant prefix. A system message, when present, is part of both $C_t$ and $U_t$.

\subsection{Context pollution}
Let $\mathcal{T}^*$ denote the true task specification, which the user reveals incrementally across $U_t$. In principle, $U_t$ gives a progressively more complete view of $\mathcal{T}^*$. However, each assistant message $a_i$ encodes the assistant's interpretation $\tilde{\mathcal{T}}_i$ of the task at turn $i$, and these interpretations may diverge from $\mathcal{T}^*$ as new user messages refine or contradict earlier ones. Conditioning on $C_t = (U_t, A_{t-1})$ then biases the model toward the stale interpretations, so that, on average over instances,
\begin{equation}
p_\theta(\text{correct} \mid U_t, A_{t-1}) < p_\theta(\text{correct} \mid U_t).
\end{equation}
The assistant's own prior outputs \emph{reduce} the probability of a correct response. Figure~\ref{fig:problem} illustrates the dynamic: an early misinterpretation persists in the context window and re-anchors the model's reasoning even after the user has provided enough information to correct it. \citet{huang2026context} confirm the inequality empirically: omitting $A_{t-1}$ (assistant omission, AO) recovers much of the lost accuracy.

\subsection{Why omission is not enough: referential vs.\ self-contained turns}
\label{sec:referential}
AO assumes that $U_t$ alone determines a correct $a_t$. We call such turns \emph{self-contained}; their counterpart is \emph{referential}, where a correct $a_t$ depends on content that exists only in $A_{t-1}$. Referential turns are common. In chat without environment state, users build on assistant-generated content: they edit a draft the assistant wrote, refine code from a previous turn, or ask a follow-up about a specific earlier step~\citep{huang2026context}. We call a referential turn \emph{stateful} when the content it depends on is environment state, as in tool use, where tool-call results are recorded in $A_{t-1}$. Omitting them makes the agent repeat actions at best and lose track of the environment at worst. In every referential case, discarding assistant turns breaks the reference. Sharded LiC~\citep{laban2025lost} sits at the self-contained end of this spectrum and stateful tool use~\citep{barres2025tau2} at the other.\footnote{The term follows the decontextualization literature~\citep{choi2021decontextualization}, which asks whether an utterance can be understood on its own. ``Stateful'' alone would conflate reference to earlier \emph{utterances} with reference to the \emph{environment}; we reserve it for the latter.}

\subsection{Editing operator desiderata}
We seek an \emph{editing operator} $\mathcal{E}$ that produces a modified context $C'_t = \mathcal{E}(C_t)$ with two properties:
\begin{enumerate}
    \item \textbf{Pollution suppression.} $C'_t$ suppresses the biasing effect of stale interpretations in $A_{t-1}$, so that $p_\theta(\text{correct} \mid C'_t) \geq p_\theta(\text{correct} \mid C_t)$.
    \item \textbf{Preservation.} $C'_t$ retains the task-relevant information in $U_t$ and $A_{t-1}$, so that $p_\theta(\text{correct} \mid C'_t) \geq p_\theta(\text{correct} \mid U_t)$ whenever $A_{t-1}$ contains state the next turn needs.
\end{enumerate}
AO satisfies the first property by construction (it replaces $A_{t-1}$ with $\emptyset$) but violates the second: it cannot retain what it discards. \method's operators do not keep assistant spans verbatim; our detector evaluation shows that they are not span-selective (Section~\ref{subsec:detector-eval}). It is designed to carry task-relevant information forward by restating it: the rebuilt context contains the specification drawn from $U_t$, the analyzer's record of the prior work it verified, and, in stateful settings, the environment state it has observed. We have not measured how much is carried forward this way, and injected correct facts mostly do not survive a reset (Section~\ref{subsubsec:tierA}).

\section{Methods}
\label{sec:methods}

We implement the editing operator $\mathcal{E}$ of Section~\ref{sec:problem-setting} as an analyzer that inspects the conversation before each assistant turn. A model cannot reliably tell its own correct work from its own mistakes when both sit in the same assistant messages, so \method externalizes that judgment: the analyzer consolidates what the user asked for from the user's messages alone, compares the assistant's work against it, and rebuilds the conversation around that specification rather than around the assistant's prior reasoning (Algorithm~\ref{alg:context-edit}, Appendix~\ref{app:algorithm}).

\subsection{Conversation analysis via subtask decomposition}
\label{sec:decomposition}

Identifying pollution requires ground truth, and in multi-turn conversation the user's messages are that ground truth: an approach is invalidated when later user messages contradict its assumptions. Because new information can look compatible with a polluted approach and users revise earlier requirements, the specification must be \emph{consolidated} from the union of user messages. Two design requirements follow. First, the specification must be \emph{grounded} in the user's requirements rather than in the assistant's interpretation, matching findings that LLMs cannot reliably self-correct without external feedback~\citep{kamoi2024criticalselfcorrect} and cannot locate their own reasoning errors, although they can correct them once told where they are~\citep{tyen2024errors}. Second, the extraction itself must be \emph{shielded} from the assistant's reasoning, or it becomes polluted too. We therefore split the analysis into two subtasks (Algorithm~\ref{alg:context-edit}, lines 2--5; tested in Section~\ref{sec:cognitive-hazard}):
\begin{itemize}[leftmargin=*, topsep=2pt, itemsep=1pt, parsep=0pt]
    \item \textbf{Subtask 1: specification extraction.} The model sees only $U_t$ (user and system messages), with $A_{t-1}$ structurally excluded, and produces a consolidated task specification $\hat{\mathcal{T}}$. The system message grounds the output format (e.g., SQL schema, function signatures). Physically excluding assistant messages implements an attentional control that prompting alone did not achieve in our tests (Appendix~\ref{app:soft-attention}).
    \item \textbf{Subtask 2: approach evaluation.} The model receives $\hat{\mathcal{T}}$, the full context $C_t$ (including assistant messages), and optionally a memory cheatsheet $M$ (Section~\ref{sec:memory}). Grounded in $\hat{\mathcal{T}}$, it identifies what the assistant got right (\texttt{aligned}) and where it diverged (\texttt{issues}).
\end{itemize}

\subsection{Editing operators}
\label{sec:strategies}

The analysis yields $(\hat{\mathcal{T}}, \texttt{aligned}, \texttt{issues})$. Three operators implement $\mathcal{E}$ with increasing modification of the context. One rule governs when they act: an operator that only adds information acts on every turn after a warm-up, and an operator that removes content acts only when the analysis finds something to remove. The analysis (Section~\ref{sec:decomposition}) is shared by the operators (on WildChat, Gated Reset uses a later revision of the Subtask~2 prompt; Section~\ref{sec:wildchat-results}); the choice of operator is a design knob whose best setting depends on the respondent and the benchmark (Table~\ref{tab:megatable}).
\begin{itemize}[leftmargin=*, topsep=2pt, itemsep=1pt, parsep=0pt]
    \item \textbf{Augment.} Append the analysis: $C'_t = C_t \oplus (\hat{\mathcal{T}}, \texttt{aligned}, \texttt{issues})$. This adds a corrective signal but leaves the polluting content in $A_{t-1}$ visible, where it may still anchor generation.
    \item \textbf{Reset.} Replace the conversation with a filled template: $C'_t = \textsc{Template}(\hat{\mathcal{T}}, \texttt{aligned})$, followed by the latest user message. Because the analysis maps to the template deterministically (no extra LLM call), this is the cheapest operator that removes content, and it does not carry injected pollution forward the way Rewrite can (Section~\ref{subsubsec:tierA}).
    \item \textbf{Rewrite.} An additional LLM call generates a restructured context: $C'_t \sim p_\theta(\cdot \mid \hat{\mathcal{T}}, \texttt{aligned}, \texttt{issues}, C_t)$. This is the most flexible operator; it trades occasional summarization loss for arbitrary restructuring, and on WildChat it beats Reset for two of four respondents (DeepSeek-V4-Flash and Kimi-K2.6; Section~\ref{sec:wildchat-results}).
    \item \textbf{When operators act.} Augment appends its analysis on every turn after a short warm-up, because it only adds information. Reset and Rewrite remove content, so they act only when the analyzer flags substantive issues (Algorithm~\ref{alg:context-edit}, lines 6--8), which on LiC happens on 94--99\% of analyzed turns. On CollabLLM, Augment and Rewrite start at the fourth user turn and Reset at the second. Every LiC, CollabLLM, and WildChat cell of Tables~\ref{tab:megatable} and~\ref{tab:wildchat} follows this rule (Appendix~\ref{app:optionA-reruns}). \emph{Gated Reset} is Reset with a turn floor (it skips the first two user turns; the first on LiC actions, WildChat, and tau2-bench) and a cap of three resets per conversation, which reduce cost and avoid disrupting on-track conversations in agentic settings where a reset can discard tool results that are expensive to re-obtain. Gated Reset is the \method-Gated-Reset row in our results.
\end{itemize}

Our default is Gated Reset, the operator we replicate most ($N{=}3$ on LiC) and the strongest or tied-strongest \method operator on tau2-bench; on text with a strong respondent and no cost constraint, Reset without the floor and cap can do better (Section~\ref{sec:wildchat-results}). For choosing among operators more generally we offer a heuristic keyed on a property observable before deployment, the strength of the full-context baseline. In stateful settings with a strong baseline, prefer the most conservative operator (Gated Reset), since there is little erroneous reasoning to remove and much tool state to keep (on tau2-bench even Gated Reset does not beat full context); where pollution is high or the respondent is weak, heavier operators have more to remove. We have not validated this heuristic prospectively.

\subsection{Optional memory}
\label{sec:memory}

We also test whether the analysis can improve with experience, using the Dynamic Cheatsheet~\citep{suzgun2025cheatsheet}: a single plain-text memory $M$, updated globally with no retrieval step, so that retrieval quality cannot confound the effect of memory. After each problem instance, a reflection prompt abstracts transferable takeaways (e.g., common pitfalls, effective consolidation strategies) into $M$, capped at 1500 words, and $M$ is supplied to Subtask~2 on later problems. Memory is an optional add-on: at our sample sizes its gains are not separable from its own run-to-run variance (Section~\ref{sec:memory-discussion}; details in Appendix~\ref{app:memory-details}).

\section{Experiments}
\label{sec:experiments}

Unless otherwise noted, GPT-5-mini serves as both the assistant and the analyzer. We call the model that generates the assistant turns the \emph{respondent}; multi-respondent experiments vary it. The analyzer is GPT-5-mini on tau2-bench and in the gpt-5.4, DeepSeek-V4-Flash, and Kimi-K2.6 columns of Table~\ref{tab:wildchat}, and each model is its own analyzer in the multi-model LiC study (Table~\ref{tab:multi-model}). The four benchmarks below differ in how much of the assistant's prior work the task needs, which lets us read how each intervention responds to increasing referentiality.

\noindent\textbf{Lost in Conversation (LiC).}
LiC~\citep{laban2025lost} shards a fully specified single-turn question into fragments that a simulated user reveals incrementally. We test four domains: math, code, database (SQL), and function calling (``actions''). We compare against AO~\citep{huang2026context}, the Concat User baseline~\citep{laban2025lost}, which joins the shards back into a single message, and ERGO~\citep{khalid2025ergo}, which resets via an LLM rewrite of prior user messages. Because each LiC instance is a single-turn question by construction, AO and Concat User are \emph{design-oracle} baselines. We run LiC in \emph{replay mode}: all strategies share the same full-context conversation prefix and only the final turn is regenerated, so comparisons measure what an intervention recovers from a fixed polluted history rather than end-to-end multi-turn behavior. Instances are difficulty-selected (Table~\ref{tab:main}); the protocol and per-task adjustments are in Appendix~\ref{app:user-sim-qc}.

\noindent\textbf{CollabLLM.}
We evaluate on MATH-Hard and BigCodeBench using the CollabLLM multi-turn framework~\citep{wu2025collabllm}. Unlike LiC, the conversation is not pre-sharded: the simulated user issues an underspecified request (still derived from a single-turn benchmark item) and both sides exchange natural follow-ups, so the assistant's prior turns contain partial work that may matter later. Evaluation details are in Appendix~\ref{app:collabllm}.

\noindent\textbf{WildChat.}
We evaluate on WildChat-1M~\citep{zhao2024wildchat} following \citet{huang2026context}. These are real human-AI dialogues in which users reference and build on earlier assistant outputs, so AO can sever the conversational thread and is no longer an oracle. Each strategy generates the next assistant turn for a real conversation prefix, and a GPT-5-mini judge compares it pairwise against AO's and full context's responses. Filtering, the two-phase setup, and the judge are in Appendix~\ref{app:wildchat-turns}; Appendix~\ref{app:trajectories} walks through a Maven/Java debugging conversation.

\noindent\textbf{tau2-bench.}
As a preliminary exploration of the agentic setting, we evaluate on the \texttt{telecom\_small} subset of tau2-bench~\citep{barres2025tau2}, in which a customer-service agent resolves a telecom troubleshooting call through diagnostic tool calls. Its dual-control design lets both the user and the agent change the environment state, so state information is spread across both parties' messages and actions. We report task reward over three seeds of 19 tasks. The agentic adaptation is in Appendix~\ref{app:agentic-methods}.

\noindent\textbf{One analyzer across benchmarks.}
All four benchmarks use the same two-subtask analysis of Section~\ref{sec:decomposition}, with documented departures, of which two are substantial. The GPT-5-mini CollabLLM result uses an earlier single-pass Rewrite over the full transcript, without structural exclusion (Section~\ref{sec:collab-results}). On tau2-bench, the analysis is re-implemented inside the tau2-bench agent with prompts adapted to tool calls: the analyzer tracks environment state and frames its output as a strategic reflection (Appendix~\ref{app:agentic-methods}). The Rewrite prompt for Table~\ref{tab:megatable} and the gpt-5.4, DeepSeek-V4-Flash, and Kimi-K2.6 columns of Table~\ref{tab:wildchat} was selected among five variants on the DeepSeek-V4-Flash LiC pool (Appendix~\ref{app:prompts}). The operators and their order of intensity are the same everywhere.

\section{Results}
\label{sec:results}

\subsection{Lost in Conversation}
\label{sec:lic-results}

\begin{table}[!htbp]
\centering
\caption{\textbf{Per-task LiC results (GPT-5-mini, accuracy \%, replay mode).} Instances are difficulty-selected (those the full-context baseline most often failed), so absolute accuracies are low by construction and are not estimates of GPT-5-mini's multi-turn accuracy. $^{\diamond}$Design oracles: single-turn by construction. $^{\ddagger}$Mean of $N{=}3$ runs; other rows are single trials. $^{\S}$Interval; see Appendix~\ref{app:tab-main-notes}, which also gives pools, denominators, and paired tests.}
\label{tab:main}

\begin{tabular}{lcccc}
\toprule
\textbf{Strategy} & \textbf{Math}~$\uparrow$ & \textbf{Code}~$\uparrow$ & \textbf{Database}~$\uparrow$ & \textbf{Actions}~$\uparrow$ \\
\midrule
\multicolumn{5}{l}{\emph{Prior work \& baselines}} \\
Baseline (full context) & \scored{60}{60.0} & \scored{15}{15.8} & \scored{4}{4.0} & \scored{34}{34.8} \\
\quad + Memory & \scored{55}{55.0} & \scored{21}{21.1} & \scored{4}{4.0} & \scored{34}{34.8} \\
AO~\citep{huang2026context}$^{\diamond}$ & \scored{85}{85.0} & \scored{73}{73.7} & \scored{32}{32.0} & \scored{82}{82.6} \\
Concat User~\citep{laban2025lost}$^{\diamond}$ & \scored{80}{80.0} & \scored{68}{68.4} & \scored{32}{32.0} & \scored{87}{87.0} \\
ERGO~\citep{khalid2025ergo}$^{\S}$ & \scored{80}{80.0} & \scored{44}{43.9} & \scored{12}{12.0} & \scored{48}{43.5--52.2} \\
\midrule
\multicolumn{5}{l}{\emph{\methodfull{} (ours)}} \\
\method-Augment & \scored{80}{80.0} & \scored{52}{52.6} & \scored{32}{32.0} & \scored{47}{47.8} \\
\quad + Memory & \scored{90}{90.0} & \scored{68}{68.4} & \scored{44}{44.0} & \scored{47}{47.8} \\
\method-Reset & \scored{75}{75.0} & \scored{57}{57.9} & \textbf{\scored{48}{48.0}} & \scored{52}{52.2} \\
\quad + Memory & \scored{85}{85.0} & \scored{68}{68.4} & \scored{44}{44.0} & \scored{52}{52.2} \\
\method-Gated-Reset$^{\ddagger}$ & \scored{80}{80.0} & \scored{64}{64.4} & \scored{39}{38.7} & \scored{61}{61.3} \\
\bottomrule
\end{tabular}
\end{table}

\begin{table}[t]
\centering
\caption{\textbf{Results across respondents on LiC, CollabLLM, and tau2-bench.} Benchmark-averaged accuracy (LiC: four tasks on a second difficulty-selected pool, Section~\ref{sec:multi-model}; CollabLLM: MATH-Hard and BigCodeBench) or task reward (tau2-bench \texttt{telecom\_small}, mean of $N{=}3$ seeds). Respondents: \mgpt, \mdsv~(DeepSeek-V4-Flash), \mkimi~(Kimi-K2.6). \method-Reset has no turn floor or reset cap; $^{*}$Gated Reset, the only Reset variant run on tau2-bench.}
\label{tab:megatable}
\footnotesize
\setlength{\tabcolsep}{4pt}
\begin{tabular}{l ccc | ccc | ccc}
\toprule
& \multicolumn{3}{c|}{\textbf{LiC}} & \multicolumn{3}{c|}{\textbf{CollabLLM}} & \multicolumn{3}{c}{\textbf{tau2-bench}} \\
\textbf{Strategy} & \mgpt & \mdsv & \mkimi & \mgpt & \mdsv & \mkimi & \mgpt & \mdsv & \mkimi \\
\midrule
Baseline                 & \scored{60}{60.4} & \scored{51}{51.3} & \scored{60}{60.4} & \scored{55}{55.0} & \scored{52}{52.5} & \scored{60}{60.0} & \scored{68}{68.4} & \scored{70}{70.2} & \scored{79}{78.9} \\
AO                       & \scored{74}{74.5} & \scored{69}{69.5} & \scored{68}{68.2} & \scored{60}{60.0} & \scored{52}{52.5} & \scored{62}{62.5} & \scored{0}{0.0}   & \scored{0}{0.0}   & \scored{0}{0.0}   \\
\method-Augment          & \scored{77}{77.0} & \scored{67}{67.0} & \scored{73}{73.8} & \scored{52}{52.5} & \scored{57}{57.5} & \scored{55}{55.0} & \scored{47}{47.4} & \scored{51}{50.9} & \scored{58}{57.9} \\
\method-Reset            & \scored{77}{77.2} & \scored{68}{68.4} & \scored{74}{74.3} & \scored{62}{62.5} & \scored{55}{55.0} & \scored{59}{59.9} & \scored{58}{57.9$^{*}$} & \scored{58}{57.9$^{*}$} & \scored{72}{71.9$^{*}$} \\
\method-Rewrite          & \scored{76}{76.6} & \scored{64}{64.2} & \scored{74}{74.5} & \scored{57}{57.5} & \scored{52}{52.5} & \scored{57}{57.5} & \scored{47}{47.4} & \scored{58}{57.9} & \scored{67}{66.7} \\
\bottomrule
\end{tabular}
\end{table}

\begin{table}[t]
\centering
\caption{\textbf{WildChat: pairwise win rates (\%) by respondent.} Each cell is the rate at which a GPT-5-mini judge prefers the \method{} response over assistant omission (AO) or full context (FC); 50\% is parity. ---: not reported. Turn sets and judge checks are in Appendix~\ref{app:wildchat-turns}.}
\label{tab:wildchat}
\footnotesize
\setlength{\tabcolsep}{4pt}
\begin{tabular}{l cc | cc | cc | cc}
\toprule
& \multicolumn{2}{c|}{\textbf{GPT-5-mini}} & \multicolumn{2}{c|}{\mgpt} & \multicolumn{2}{c|}{\mdsv} & \multicolumn{2}{c}{\mkimi} \\
\textbf{Strategy} & vs AO & vs FC & vs AO & vs FC & vs AO & vs FC & vs AO & vs FC \\
\midrule
\method-Augment      & ---              & ---              & \scored{86}{86.0} & \scored{78}{78.9} & \scored{89}{89.5} & ---              & \scored{82}{82.9} & ---              \\
\method-Reset        & \scored{83}{83.0} & \scored{83}{83.5} & \scored{88}{88.6} & \scored{77}{77.3} & \scored{75}{75.0} & ---              & \scored{71}{71.6} & ---              \\
\method-Gated-Reset  & \scored{86}{86.1} & \scored{83}{83.8} & \scored{74}{74.1} & \scored{72}{72.4} & ---              & ---              & ---              & ---              \\
\method-Rewrite      & \scored{82}{82.6} & \scored{80}{80.3} & \scored{83}{83.3} & \scored{72}{72.9} & \scored{79}{79.2} & \scored{73}{73.6} & \scored{91}{91.5} & \scored{76}{76.3} \\
\bottomrule
\end{tabular}
\end{table}

\noindent\textbf{\method closes most of the gap to the design oracle.}
On the Gated Reset row (mean of $N{=}3$ replay-mode reruns; Table~\ref{tab:main}), \method reaches 80.0\% on math, closing 80\% of the 25.0pp gap between full context and AO, and 64.4\% on code, closing 84\% of the 57.9pp gap; on math, code, and actions the closure ranges from 55\% (actions, where Gated Reset uses raw $n{=}25$ against the other rows' 23) to 84\% (code). On database, both Gated Reset (38.7\%) and Reset (48.0\%, a single trial) exceed the design oracles (32.0\%), possibly because consolidating the shards into a schema-grounded specification helps more than removing assistant turns alone. On math the gain is not separable from the analyzer solving the problem: on the three-respondent matrix 47\% of math analyzer outputs contain the answer, and on the rest the paired gain is $-2.6$pp ($p{=}0.815$); on code and database (0--1\% leakage) it holds on leak-free items ($+30.2$pp, $+26.0$pp). The Gated Reset reruns have standard deviations of 5.0--7.2pp, and their means clear the full-context baseline on every LiC task and match or exceed the no-memory Augment operator (Appendix~\ref{app:variance}).

\noindent\textbf{The gain holds on unselected pools.}
Difficulty selection can inflate gains through regression to the mean, so we also ran LiC database end-to-end on the complete, unselected pool ($n{=}107$; assistant gpt-5.4-mini; Section~\ref{subsec:condenser-mtosc}). There, full context scores 56.1\%, AO 69.2\%, Concat User 63.6\%, and Reset 75.7\%, above both design oracles; on the unselected code pool ($n{=}100$), Reset scores 92.0\% against full context's 83.0\% ($+9.0$pp, exact McNemar $p{=}0.023$).

\noindent\textbf{Removing polluted content helps beyond the analysis.}
With the same analysis, Reset and Gated Reset score above Augment on code (+5.3--11.8pp) and database (+6.7--16.0pp). The operators differ only in whether the assistant still sees the polluted history, which is consistent with anchoring on that history being an effect separate from the fragmentation of the task across shards. Except for Gated Reset, these comparisons rest on single trials.

\noindent\textbf{Statistics and scale.}
On the three-respondent LiC matrix of Table~\ref{tab:megatable} (113--150 conversations per respondent and task), Reset improves over full context in $33$ of $36$ paired comparisons (3 respondents $\times$ 4 tasks $\times$ 3 replay prefixes; mean $+15.9$pp, sign test $p<0.0001$). This is a single fixed configuration, not a per-cell maximum, and a paired effect on a failure-selected pool (its replay prefixes are also weighted toward baseline failures), not a population estimate. At the item level, Reset gains $+15.4$pp (McNemar $p<10^{-4}$; $350$ item-level wins to $93$) with a problem-clustered $95\%$ interval of $[+11.5,+19.4]$. We report this figure as a descriptive aggregate: the $1{,}668$ items are $191$ problems observed under up to $9$ correlated replicates each, so they are not independent, and the interval comes from a problem-clustered bootstrap rather than from treating items as independent trials.

\noindent\textbf{Generalization across model families.}
\label{sec:multi-model}
Alongside GPT-5-mini, we evaluate three additional models (GPT-5; DeepSeek V3.2, open-weight with 37B active parameters; and Qwen 3.5 35B-A3B, with 3B active) on a harder subset of 20 instances per task selected from GPT-5.2 logs (Appendix~\ref{app:multi-model-protocol}). Reset improves accuracy in 15 of 16 model-task pairs (Table~\ref{tab:multi-model}), by +20 to +42pp on average per model. Table~\ref{tab:megatable} adds three further respondents (gpt-5.4, DeepSeek-V4-Flash, Kimi-K2.6) on the second GPT-5.2-selected pool (50 instances per task). The best \method operator beats full context in every LiC cell, and the three operators land within about 4pp of each other; CollabLLM and tau2-bench are discussed in Sections~\ref{sec:collab-results} and~\ref{sec:tau2-results}. No single operator dominates everywhere.

\subsection{Matched-compute baselines: faithful condensation and MT-OSC}
\label{subsec:condenser-mtosc}

A natural objection is that \method{}'s gains come from extra compute or from compression per se rather than from adjudicating polluted reasoning. Two baselines on LiC argue against this. First, a per-turn LLM \emph{condenser} instructed to compress faithfully (to preserve content, not to find errors) does not close the gap on the full, unselected pool. On LiC-database (assistant gpt-5.4-mini, full context $56.1\%$), the condenser scores $53.3\%$ at one call per turn ($-2.8$pp, $p=0.68$) and $47.7\%$ at two calls per turn ($-8.4$pp, $p=0.08$); a replicate of the one-call arm also scored $47.7\%$, so the condenser is neutral to negative at either budget rather than degrading with budget, and a neutral-phrasing variant reproduces the result at $51.4\%$. \method{}-Reset reaches $75.7\%$ ($+19.6$pp, $p=0.0005$) on the same conversations, and the pattern repeats on code ($79.0$/$80.0\%$ for the condenser vs.\ $83.0\%$ for full context and $92.0\%$ for Reset, $+9.0$pp, $p=0.023$). Compute does not explain the difference: the two-call condenser used more than Reset ($1.02$--$1.19\times$ its strategy calls, $1.62$--$2.14\times$ its tokens) and still lost, while Gated Reset reaches $+17.8$pp at $0.41\times$ Reset's calls. The mechanism is visible directly: with the error-finding clause removed, the condenser flags an assistant error in $0$ of $340$ summaries, so faithful summarization carries invalidated reasoning forward rather than adjudicating it.

Second, we reimplemented MT-OSC~\citep{singh2026mtoscpathllmslost}, a multi-turn one-off sequential condensation method that condenses the chat history in the background (prompt, exemplars, recursion, and hyperparameters verbatim, as no code is released; condenser gpt-5.4-mini instead of Llama-3.3-70B, matching our other arms). At its published window $w{=}4$ on LiC-database it is a near no-op: $60.7\%$ vs.\ $56.1\%$ ($+4.7$pp, $p=0.383$). It edits the context in only $6$ of $107$ conversations, so the gain comes almost entirely from conversations it never touches; forcing engagement at $w{=}2$ makes it worse ($47.7\%$). Length-triggered schedules cannot reach the early pollution that LiC produces. We report this as scoped evidence on short conversations, not a verdict on MT-OSC's long-horizon regime. Both baselines were run on LiC only.

\subsection{CollabLLM}
\label{sec:collab-results}

CollabLLM sits near the middle of the referentiality spectrum, and its results are mixed. The GPT-5-mini run uses an earlier single-pass Rewrite over the full transcript, without structural exclusion; the three-respondent runs of Table~\ref{tab:megatable} use the standard analysis and the operator rule of Section~\ref{sec:strategies}. Averaged over two GPT-5-mini runs, Rewrite is below full context on MATH-Hard (45.0 vs.\ 50.0) and above it on BigCodeBench (71.2 vs.\ 67.5, scored by a conversation-aware GPT-5 judge), where AO scores higher still (85.0); one of the two runs alone was above full context on both tasks (Appendix~\ref{app:collabllm}). Across the three additional respondents (Table~\ref{tab:megatable}), Reset, the one operator that acts in most CollabLLM conversations, scores 62.5 against AO's 60.0 and full context's 55.0 on gpt-5.4, 55.0 against 52.5 and 52.5 on DeepSeek-V4-Flash, and 59.9 against 62.5 and 60.0 on Kimi-K2.6. Each of these gaps is at most three problems out of 40. Augment and Rewrite act only from the fourth user turn, which few conversations reach (8 and 12 of 120), so their rows mostly measure full context under fresh sampling; this includes DeepSeek-V4-Flash's highest row (Augment, 57.5), which acted in 3 of its 40 conversations. The clearest CollabLLM evidence is an execution-scored three-replicate run with DeepSeek-V4-Flash, in which Reset beats full context on BigCodeBench by $+15$pp in every replicate.

\subsection{WildChat}
\label{sec:wildchat-results}

Unlike LiC, WildChat contains many referential turns, so AO is no longer an oracle and can degrade quality. Following \citet{huang2026context}, we classify each turn as \texttt{new\_ask}, \texttt{feedback}, or \texttt{no\_feedback} (the user continues without explicitly referencing the assistant's response, but often builds on it implicitly; definitions in Appendix~\ref{app:wildchat-turns}). Phase~1 of the protocol reproduces their finding: AO helps on \texttt{new\_ask} and \texttt{feedback} turns but \emph{hurts} on referential \texttt{no\_feedback} turns, where full context wins the pairwise comparison.

\noindent\textbf{\method outperforms both AO and full context.}
On GPT-5-mini, Gated Reset achieves an 86.1\% win rate vs.\ AO and 83.8\% vs.\ full context while intervening on 72.3\% of turns (Table~\ref{tab:wildchat}); on the \texttt{no\_feedback} turns, where AO is weakest, it still wins 85.1\% vs.\ AO and 80.6\% vs.\ full context, and its gate opens less often there (Appendix~\ref{app:wildchat-turns}). Every printed cell in Table~\ref{tab:wildchat} lies between 71.6\% and 91.5\%. All of these are preferences of a single GPT-5-mini judge; their sensitivity to the judge is discussed in Section~\ref{sec:discussion}. With DeepSeek V3.2 as respondent, Reset wins 72.5\% vs.\ AO and 72.5\% vs.\ full context (Table~\ref{tab:wildchat-memory}). Appendix~\ref{app:trajectories} walks through a Maven-debugging conversation in which Gated Reset wins all 14 comparisons across 7 turns.

\noindent\textbf{Reset with and without a floor and cap on a strong respondent.}
By design, the two Reset variants differ only in the turn floor and the reset cap, since both act only when the analyzer flags an issue. On LiC they coincide: replay regenerates only the final turn, which is past the floor, and the cap never binds. They diverge on WildChat with gpt-5.4: Reset wins 88.6\% vs.\ AO and Gated Reset 74.1\%. On the 35 prefixes scored in both arms (of 44 and 58), the gap is $-$14.3pp but rests on seven discordant turns (6 vs.\ 1, exact McNemar $p{=}0.125$), so we read it as a direction. As run on WildChat, however, the two arms also differ in the Subtask~2 prompt revision and the reset template, so the gap cannot be attributed to gating alone.

\subsection{tau2-bench}
\label{sec:tau2-results}

AO collapses to \textbf{0\%} on tau2-bench for every respondent we evaluated (Table~\ref{tab:megatable}, tau2 block). Stripping assistant messages erases every diagnostic result the agent has fetched, so the agent repeats its lookups until the step budget ends the episode; omission is incompatible with stateful tool use in this setting. The collapse holds at $N{=}3$ across 9 cells and 171 rollouts (170 scored) without a single non-zero reward, through a confirmed step-budget (\texttt{max\_steps}) termination path, and is significant against full context for every respondent (paired $-$68.4 / $-$69.6 / $-$78.9pp over completed rollouts, $p<0.0001$; one DeepSeek-V4-Flash AO rollout errored).

\noindent\textbf{No \method operator improves over full context.} Across the three-respondent matrix at $N{=}3$ (855 rollouts, one of which errored), full context scores $68.4 \pm 13.9$ (gpt-5.4), $70.2 \pm 11.0$ (DeepSeek-V4-Flash), and $78.9 \pm 0.0$ (Kimi-K2.6), and no \method operator improves on it. Among the \method operators, Gated Reset is the strongest or tied-strongest on all three respondents ($57.9 \pm 21.1$ / $57.9 \pm 10.5$ / $71.9 \pm 11.0$) and stays within roughly 7 to 12pp of full context (all $p>0.14$), while Augment is significantly worse on all three. Single runs on this benchmark can be unrepresentative: in an earlier single run on gpt-5.4, Augment and Rewrite scored 84.2 and 73.7, against 47.4 for each at $N{=}3$, while full context and Gated Reset gave similar values in both. We found no cause (we ruled out model substitution, gating failure, degenerate termination, and rate limits), which is one reason we report only $N{=}3$ means.

On GPT-5-mini (Appendix~\ref{app:tau2-diagnostic}), Gated Reset likewise stays within trial noise of full context across 3 seeds. Across all four respondents the load-bearing finding is the same: AO collapses whenever tool-call results live only in assistant turns, while \method does not collapse, although it does not exceed full context.

\subsection{Pollution is contagious}
\label{sec:analysis-ablations}
\label{sec:cognitive-hazard}

\begin{table}[t]
\centering
\small
\begin{tabular}{lccc}
\toprule
\textbf{Configuration} & \textbf{Math} & \textbf{Code} & \textbf{Database} \\
\midrule
\method-Augment & \textbf{80\%} & \textbf{53\%} & \textbf{32\%} \\
\quad $-$ structural exclusion & 40\% & 11\% & 8\% \\
\quad\quad $-$ subtask decomposition & 55\% & 21\% & 4\% \\
\midrule
Full context (no analysis) & 60\% & 16\% & 4\% \\
\bottomrule
\end{tabular}
\caption{Progressive ablation of the analyzer (Augment, no memory; single trials on the Table~\ref{tab:main} pools). Row~2 lets the specification extractor see assistant messages; row~3 also collapses the two subtasks into a single pass.}
\label{tab:cognitive-hazard}
\end{table}

Letting the specification extractor see assistant messages erases most of the analyzer's gain (Table~\ref{tab:cognitive-hazard}: 80\%$\to$40\% on math, 53\%$\to$11\% on code, 32\%$\to$8\% on database; single trials), to or below the full-context level on math and code. Even a model explicitly tasked as an independent reviewer anchors on the assistant's flawed reasoning. The polluted single-pass variant then \emph{beats} the polluted two-step one on math and code (55\% vs.\ 40\% on math), consistent with the second subtask treating the biased specification as ground truth. Decomposition helps only when earlier stages produce clean outputs, a structural data-flow constraint that prompting alone does not enforce (Appendix~\ref{app:hard-attention}).

\subsection{The analyzer as a pollution detector}
\label{subsec:detector-eval}

To test whether the analyzer identifies harmful context rather than editing arbitrarily, we evaluate it directly as a detector, with no judge in the label path so that the evaluation cannot be circular. We report two studies in full, including the results that are unfavorable to the method.

\subsubsection{Tier A: constructed injection, ground-truth labels}
\label{subsubsec:tierA}

Into $145$ LiC database and code conversations we inject two assistant-side spans, one known-false and one known-true, in an identical frame, so labels are correct by construction. Four offline controls (identity editor $0.00$ removal; hand-removal oracle $1.00$; two delete-everything editors $1.00$ removal, $0.00$ preservation) show that preservation, not removal, is the metric that cannot be gamed. On the probe-admissible set ($n{=}126$ conversations that pass a mechanical check that the span probe is valid), \method{}-Reset removes the injected pollutant $97.6\%$ of the time ($123/126$) and names it explicitly in its \texttt{issues} output $78.6\%$ of the time ($89.7\%$ on the causally harmful subset), at a gate sensitivity of $98.4\%$ (against a $96.8\%$ gate-open rate on the clean arm, so the gate itself barely discriminates); the injected true span is also named as an issue $42.1\%$ of the time. Detection of the pollutant is reliable, but selectivity is absent: preservation is $4.0\%$ and edit precision sits at chance, because Reset discards the assistant side wholesale. Rewrite removes only $27.0\%$ here, so the metric is not saturated by construction. A factorial on the causally validated harmful subset ($n{=}97$) shows the accuracy consequence: full context scores $24.7\%$ clean, drops to $9.3\%$ with the injected pollutant present, and recovers to $59.8\%$ under \method{}-Reset with that pollutant still in the history.

\subsubsection{Tier B: causal span ablation, no LLM in the label path}
\label{subsubsec:tierB}

Constructed spans may be more salient than natural pollution, so we also run a counterfactual study. For $111$ spans occurring naturally in $30$ LiC conversations (the assistant's own prose and code, nothing injected), we re-ran the final turn with each span present and with it removed, byte-identical otherwise ($3{,}357$ assistant turns, $0$ errors); a span is causally harmful if removing it raises accuracy. The label is a measured accuracy difference, with no detector, judge, or LLM anywhere in the label path, and three controls pass in both directions (a contentless span $+0.033$, n.s.; the injected pollutant $+0.368$; a full-specification-plus-gold-SQL span $-0.447$). Natural pollution is real and concentrated: the spread of per-span effects exceeds a replicate-matched null (SD $0.155$ vs.\ $0.125$, $p=0.0085$; $16$ spans with $|\Delta|\!\geq\!0.25$ where the null predicts $9.3$, $p=0.017$), while the mean span is inert ($+0.020$). Pollution is carried by a minority of spans rather than spread evenly over the history. \method{} removes all $7$ probe-admissible spans labelled harmful ($100\%$), as expected of an operator that discards most assistant content.

We are explicit about what this does \emph{not} show. Neither operator is span-selective: Reset keeps only $5$ of $66$ probe-admissible spans and Rewrite $0$ of $66$, and preservation of causally useful spans is $0\%$. A label-free test agrees: the mean effect of the spans Reset removed minus those it kept is $-0.014$ (permutation $p=0.85$), where a selective editor would score positive. The correct statement of the mechanism is therefore not surgical span preservation but \emph{detect, discard the assistant side, and rebuild the specification from the user side}; the accuracy gain comes from that rebuild. This also fixes the scope of the method's safety: structural exclusion is safe exactly where user turns independently specify the task. Where they do not, referents can reach the rebuilt context only through Subtask~2's \texttt{aligned} record, which does see the assistant's turns.

\subsubsection{Analyzer robustness across model families}
\label{subsec:analyzer-robustness}
Because the analyzer is the one component shared by every operator, we swap it across five models from four families while holding the respondent and operator fixed (DeepSeek-V4-Flash, Reset, LiC code and database). Every analyzer produces a positive, individually significant gain over full context, from $+12.9$pp (gpt-4o-mini) to $+39.9$pp (Kimi-K2.6), and the ordering tracks capability rather than provenance: weak analyzers under-detect rather than mis-detect (details in Appendix~\ref{app:analyzer-robustness}).

\section{Discussion}
\label{sec:discussion}

\paragraph{One analyzer, three operators.}
The WildChat results (Table~\ref{tab:wildchat}) show that the best operator depends on the respondent while the analysis feeding the operators is shared. With Kimi-K2.6, Rewrite wins 91.5\% vs.\ AO, ahead of Augment (82.9\%) and Reset (71.6\%), plausibly because Reset's template strips conversational structure the next turn refers to; with gpt-5.4 the order reverses (Reset 88.6\%, Rewrite 83.3\%), possibly because Reset passes the analyzer's specification through without a second LLM call. On tau2-bench no operator beats full context and Augment is significantly worse than Gated Reset's parity, so we read no direction for operator intensity from the agentic setting beyond preferring the conservative operator. We therefore present \method as decoupling the analysis (one shared analyzer) from the intervention intensity (a per-deployment knob), rather than as one operator that is best everywhere; Section~\ref{sec:strategies} gives our default.

\paragraph{Memory is a direction, not a measured effect.}
\label{sec:memory-discussion}
The optional memory (Section~\ref{sec:memory}) is not a measured effect. Its single-trial LiC rows sit above Augment on math, code, and database (Table~\ref{tab:main}), and on WildChat it improves DeepSeek V3.2's Reset by +13.8pp but not near-ceiling GPT-5-mini ($-$4.1pp; Table~\ref{tab:wildchat-memory}). But a remeasurement of the database cell over four trajectory orderings (gpt-5.4-mini respondent) found memory $8.0$pp \emph{below} no memory on average, with a standard deviation of roughly 6pp. Any gain is not explained by leakage: the learn set does not overlap the LiC evaluation set, and a direct test of the transductive online protocol found no effect (Appendix~\ref{app:memory-details}).

\paragraph{Limitations.}
Our LiC results replay difficulty-selected instances, so they measure recovery from a fixed polluted history on hard items; the unselected-pool results (Section~\ref{sec:lic-results}) are the exception. The WildChat preference is that of a GPT-5-mini judge: one cross-family check agrees with it on 87.5\% of pairs, but in another check three other judges preferred AO over Reset on the same responses, and with GPT-5 as a self-judged respondent \method did not beat AO or full context (Appendix~\ref{app:wildchat-turns}). In the fully simulated referential benchmarks (CollabLLM, tau2-bench) \method does not reliably beat full context. LiC, CollabLLM, and tau2-bench use simulated users, and most LiC, CollabLLM, and WildChat cells are single trials; our LiC and CollabLLM replicates vary only temperature-1.0 sampling for a fixed history, not the user simulator. \method costs two to three extra LLM calls per analyzed turn (Appendix~\ref{app:eval-details}); folding the analysis into the base model by self-distillation~\citep{zhao2026selfdistillreason,hubotter2026rlselfdistill,shenfeld2026selfdistill} is untested. The instantiation differs in documented ways across benchmarks (Section~\ref{sec:experiments}), and we have not measured specification extraction on turns whose referent lives only in assistant turns. Long-horizon tasks and broader agentic domains remain untested.

\section{Related work}
\label{sec:related}

\noindent\textbf{Multi-turn degradation and context pollution.}
\citet{laban2025lost} introduced LiC; \citet{huang2026context} named the failure mode \emph{context pollution} and showed that AO recovers much of the lost performance; ERGO~\citep{khalid2025ergo} resets the conversation through an LLM rewrite of prior user turns when output entropy spikes. These interventions decide \emph{when} to act, but once triggered they remove or rewrite context wholesale, and none audits the assistant's prior work against a user-grounded specification. Wholesale removal works where turns are self-contained (the LiC setting, where AO is a design oracle); on referential turns omission degrades quality, and in stateful tool use it collapses (Section~\ref{sec:tau2-results}).

\noindent\textbf{Agentic context management.}
Concurrent work on long-horizon tasks (THREAD~\citep{schroeder2025thread}, Recursive Language Models~\citep{zhang2025rlm}, Context Folding~\citep{sun2025contextfolding}, U-Fold~\citep{su2026ufold}, MemoBrain~\citep{qian2026memobrain}) targets \emph{compaction}, fitting more reasoning into a bounded context; we address \emph{context pollution}, where prior outputs mislead generation even within ample context. System 2 Attention~\citep{weston2023s2a} regenerates the context to strip biasing content before answering, in a single call and without auditing prior work against a user-only specification. MT-OSC~\citep{singh2026mtoscpathllmslost}, which condenses multi-turn chat history, is closest in setting, and we compare against it in Section~\ref{subsec:condenser-mtosc}. Detailed comparisons and additional related work are in Appendices~\ref{app:additional-related}--\ref{app:agentic-context}.

\section{Conclusion}
\label{sec:conclusion}

We proposed \methodfull{} (\method), a training-free method in which an analyzer, shielded from the assistant's turns, rebuilds the task specification from the user side and edits the context around it. On difficulty-selected LiC instances it closes 55--84\% of the gap to the design oracle on three tasks and exceeds it on the fourth; a GPT-5-mini judge prefers it over full context and AO in every configuration of Table~\ref{tab:wildchat}; CollabLLM is mixed; and on tau2-bench, where AO collapses to 0\%, \method does not collapse but does not exceed full context. Its operators are not span-selective, so the gain comes from the rebuild.

Two lessons follow. First, pollution is contagious: even a separate reviewer model anchors on the assistant's reasoning if it can see it, and in our experiments only structural exclusion, not prompting, prevented this. Second, the analyzer is shared but the best operator depends on the respondent, a per-deployment choice. Broader end-to-end, long-horizon, and real-user tests come next.

\bibliographystyle{plainnat}
\bibliography{neurips_2026_conference}

\appendix

\section{Algorithm}
\label{app:algorithm}

\begin{algorithm}[h]
\caption{\methodfull{} (\method)}
\label{alg:context-edit}
\begin{algorithmic}[1]
\REQUIRE Conversation context $C_t = (u_1, a_1, \ldots, u_t)$, operator $o \in \{\text{Augment}, \text{Reset}, \text{Rewrite}\}$
\STATE \textbf{// Subtask 1: Specification extraction}
\STATE $U_t \leftarrow (u_1, \ldots, u_t)$ \hfill $\triangleright$ Structurally exclude $A_{t-1}$
\STATE $\hat{\mathcal{T}} \leftarrow \textsc{Analyze}(U_t)$ \hfill $\triangleright$ Extract clean task specification
\STATE \textbf{// Subtask 2: Approach evaluation}
\STATE $(\texttt{aligned}, \texttt{issues}) \leftarrow \textsc{Evaluate}(\hat{\mathcal{T}}, C_t)$
\IF{$o \in \{\text{Reset}, \text{Rewrite}\}$ \AND $\texttt{issues} = \emptyset$}
    \RETURN $C_t$ \hfill $\triangleright$ No edit needed (pass-through)
\ENDIF
\STATE \textbf{// Apply editing operator $\mathcal{E}$}
\STATE $C'_t \leftarrow \textsc{Intervene}(\hat{\mathcal{T}}, \texttt{aligned}, \texttt{issues}, C_t)$
\RETURN $C'_t$
\end{algorithmic}
\end{algorithm}

\section{Evaluation details}
\label{app:eval-details}

\subsection{LiC evaluation adjustments}
\label{app:user-sim-qc}

We make task-specific adjustments to the LiC evaluation protocol to reduce false negatives caused by overly strict extraction and ambiguous formatting, and apply a post-hoc false negative identification procedure across all tasks.

\paragraph{Replay-mode protocol.} To isolate intervention effects, we run all LiC strategies in \emph{replay mode}: every strategy shares the same baseline conversation prefix and only the final turns are regenerated under each strategy. Cell-to-cell comparisons therefore measure what an intervention recovers from a fixed history rather than confounding strategy effects with divergent rollouts.

\paragraph{Sampling parameters.} All LLM calls (assistant, user-simulator, system-agent, and analyzer) use \texttt{temperature}~$=$~$1.0$ with default top-$p$. This is the reasoning-model default: GPT-5-family models silently override any \texttt{temperature=0.0} request to \texttt{1.0}, so we use \texttt{1.0} throughout for consistency. We do not pass a \texttt{reasoning\_effort} parameter for analyzer calls (constraining reasoning effort severely degrades analyzer quality on GPT-5-mini in our pilots). The same sampling configuration is used both when generating the original baseline trajectory and when each strategy regenerates the final turn(s) under replay mode.

\paragraph{API cost.} All experiments are inference-only and were run via commercial APIs (OpenAI, including Azure OpenAI) and hosted endpoints for open-weight and third-party models (Azure AI Foundry for DeepSeek-V4-Flash and Kimi-K2.6; other providers for DeepSeek V3.2, Qwen, and Llama). We report per-run costs rather than totals so the figures translate directly to reproduction budget. With \texttt{gpt-5-mini} as both assistant and analyzer, a single Gated Reset replay-mode rerun of an LiC task ($n{=}18$--$25$ samples) costs approximately \$0.04--\$0.06; tau2-bench is more expensive per task because conversations are longer and tool-call payloads inflate token usage, at roughly \$0.70 per trial of the 20-task \texttt{telecom\_small} subset (19 scored tasks in the multi-respondent runs), with the Gated Reset variant adding $\sim$+20\% over the Baseline due to the analyzer's 2--3 extra LLM calls per turn. WildChat is scored by GPT-5-mini pairwise judge calls and the CollabLLM GPT-5-mini BigCodeBench results by a GPT-5 conversation judge, over pre-generated assistant responses, so per-experiment cost is dominated by judge tokens rather than full multi-turn simulation; a single judged WildChat condition over $n{\approx}175$ turns costs a few dollars. Despite our scaled intervention adding 2--3 LLM calls per turn, our project-level total is well under the $\sim$\$5{,}000 envelope reported by~\citet{laban2025lost}: \citeauthor{laban2025lost} ran six task domains across fifteen models with full multi-turn simulation, while our headline tables span four LiC tasks (with replay mode amortizing the baseline trajectories), three CollabLLM/WildChat conditions, one tau2-bench subset, and several model families.

\subsubsection{Task-specific prompt and extraction changes}
\label{app:eval-adjustments}

\paragraph{Math.} The original LiC math evaluator uses symbolic answer extraction that is overly strict, producing frequent false negatives when the assistant's final answer is correct but formatted unexpectedly. Since GSM8K problems have integer-only solutions, we simplify extraction by requiring a standardized answer prefix. Our v2 system prompt adds the instruction \texttt{**ANSWER: <number>**}, making extraction a simple regex match on a known prefix. This eliminates false negatives from formatting variation while remaining faithful to the original task.

\paragraph{Code.} The original HumanEval system prompt does not specify how the assistant should structure its code output. In the multi-turn setting, the assistant sometimes omits the function definition entirely (inlining the logic), or assumes input comes from \texttt{stdin} rather than function parameters, because the sharded problem description does not always convey the expected function signature. Our v2 system prompt explicitly requires: (1)~exactly one top-level function definition, (2)~inputs as function arguments rather than \texttt{stdin}, and (3)~a markdown Python code fence for the final answer. We also relaxed the extraction code to handle minor formatting variations (e.g., extra whitespace, trailing comments).

\paragraph{Database.} Similar to code, the original database prompt does not specify an output format, making SQL extraction fragile. Our v2 prompt requires the final query in a \verb|```sql| markdown code fence, with the constraint that only one such fence should be used for the final answer.

\paragraph{Actions (function calling).} This is the only LiC task that expects multiple output artifacts (a list of function calls). Because shards are revealed incrementally, the assistant sometimes reasonably assumes it does not need to repeat previously established function calls in its final response, but evaluation only considers the last assistant message. Our v2 system prompt adds an explicit instruction to output all function calls needed to fulfill the user's complete request at each turn, ensuring the evaluated response contains the full set.

\subsubsection{False negative identification}
\label{app:false-neg}

Through manual trace analysis, we found that the user simulator model (which converts problem shards into natural-language user messages) sometimes rephrases shards in ways that subtly distort the original meaning, introducing ambiguity, altering constraints, or omitting critical details. While LiC controls for the \emph{sharding} process by verifying that concatenated-shard performance approximates single-turn performance, it does not control for the user simulator's \emph{rephrasing} and \emph{ordering} of those shards.

We account for this with a pool-level pre-filter, computed once and applied identically to every arm. For each instance where the \emph{full-context baseline} assistant fails, we collate the user-role messages of that baseline trace and compare them against the original single-turn question. We query GPT-5 (Prompt~\ref{prompt:false-neg}, Appendix~\ref{app:prompts}) to determine whether those messages, taken as a union, provide sufficient information to solve the original problem. The resulting per-task list of instance IDs is stored alongside the replay pool and every strategy is evaluated on the same reduced pool, which is why all rows of Table~\ref{tab:main} share the denominators $n{=}20/19/25/23$ (math $23{-}3$, code $25{-}6$, database $25{-}0$; the actions column is normalized to a common 23 samples, except the Gated Reset row, which uses raw $n{=}25$; Appendix~\ref{app:variance}).

We deliberately do \emph{not} recompute this classifier per run. A reset or rewrite arm stores the \emph{edited} context, in which the original user turns have been consolidated into a single specification message; the classifier would then see about one user turn per sample on those arms against about five on the full-context baseline, and would exclude their failures at a far higher rate. Computing the filter once on baseline traces removes that asymmetry by construction: it is a property of the \emph{instance}, not of the arm.

Instances flagged as insufficient are excluded from the evaluation pool for \emph{all} arms, as the failure is attributable to the evaluation setup rather than to context pollution or the intervention strategy. All accuracies in Table~\ref{tab:main} and the main text are raw accuracies on that common pool, with no per-run adjustment (the exploratory soft-attention splits of Table~\ref{tab:soft-attention} are the one exception, noted in its caption). Manual analysis of the classifier's outputs finds it to be higher-precision, lower-recall: it conservatively flags only clear cases of information loss or distortion. We consider this acceptable because we prefer to allow some ambiguity to the assistant (counting borderline cases as valid) rather than overly invalidating results.

\subsubsection{Variance of the default operator}
\label{app:variance}

We do not report error bars on every cell of Table~\ref{tab:main} (see Limitations), but we did spot-check that the headline LiC ordering is not a lucky-run artifact. We re-ran the default operator (Gated Reset) two additional times across all four LiC tasks, giving $N{=}3$ per cell when combined with the headline run. We focus the variance budget on a single complete row rather than scattered cells, both so the row is coherent and because Gated Reset is our default operator. Two additional implicit checks come for free from the experiment design: (i)~each of \method's operator variants (Augment, Reset, Gated Reset, plus their +Memory counterparts) reuses the same upstream analyzer artifact per question, so the four-row ordering across operators is also a check on robustness across different operationalizations of the same analysis step, and Gated Reset clears the full-context baseline on every LiC task in this row, mirroring the variance result; (ii)~the multi-model results (Table~\ref{tab:multi-model}) replay the same protocol across four models from three families and recover the same qualitative ordering in 15 of 16 cells, providing a cross-model robustness check on the analysis step itself.

\paragraph{Protocol.} We use replay mode (last-turn regeneration with fixed user-sim trajectory loaded from saved full-context baseline traces). This isolates analyzer + assistant variance (the user simulator's shard-reveal trajectory is held constant across the three runs) and so likely \emph{underestimates} full end-to-end variance, which would also include user-sim stochasticity. The configuration matches the headline run for each cell, replaying the same saved baseline traces. On actions, the rebuilt context also restates the system prompt's instruction to repeat all function calls in the final turn; the other rows receive that instruction only in the system prompt. All runs use \texttt{gpt-5-mini} as both assistant and analyzer at temperature~1.0.

\paragraph{Results.} Per-run accuracies and aggregate statistics are below (raw $n$, no auxiliary normalization):

\begin{center}
\begin{tabular}{lccccc}
\toprule
\textbf{Task} & \textbf{Headline} & \textbf{Run 2} & \textbf{Run 3} & \textbf{Mean} & \textbf{Std} \\
\midrule
math (n=20)     & 75.0 & 80.0 & 85.0 & \textbf{80.0} & 5.0 \\
code (n=18--19) & 72.2 & 63.2 & 57.9 & \textbf{64.4} & 7.2 \\
database (n=25) & 44.0 & 32.0 & 40.0 & \textbf{38.7} & 6.1 \\
actions (n=25)  & 68.0 & 56.0 & 60.0 & \textbf{61.3} & 6.1 \\
\bottomrule
\end{tabular}
\end{center}

\paragraph{Reading.} Std is in the 5--7pp range. Means clear the full-context baseline (60.0/15.8/4.0/34.8) on every task and match or exceed the Augment operator without memory (80.0/52.6/32.0/47.8): $80.0 \pm 5.0$ vs Augment 80.0 (parity, expected, since math is the easy direction); $64.4 \pm 7.2$ vs Augment 52.6; $38.7 \pm 6.1$ vs Augment 32.0; $61.3 \pm 6.1$ vs Augment 47.8. The qualitative ordering (Gated Reset substantially above the full-context baseline; Reset/Gated Reset above Augment on context-removal-sensitive tasks, code/database/actions) is preserved. Specific point-estimate gap-closure percentages cited in the body (e.g., ``84\% of the gap on code'') should be read as derived from these means.

\paragraph{What this does and does not establish.} This is variance under fixed user-sim trajectories with the analyzer and assistant resampled. It does not estimate variance from re-rolling the user simulator (which would require full simulation, not replay). Reruns of the other ours-rows (Augment, Reset, +Memory) are left to future work; the choice to anchor variance evidence on Gated Reset reflects that it is our default operator (Section~\ref{sec:strategies}).
\subsubsection{Notes on Table~\ref{tab:main}}
\label{app:tab-main-notes}

\paragraph{Pools and denominators.} The full LiC pools have 103 (math), 100 (code), 107 (database), and 105 (actions) instances. Difficulty selection keeps instances that the full-context baseline answered incorrectly in at least 60\% of five independent baseline runs (four for code) and takes the 25 with the highest error rate (math had only 23 eligible instances), so gap-closure percentages are measured against a same-pool ceiling rather than against 100\%. The shared pre-filter (Appendix~\ref{app:false-neg}) then removes 3, 6, and 0 instances from math, code, and database ($n{=}23{-}3$, $25{-}6$, $25{-}0$). Actions has no filter artifact: its 23 is a legacy ``common 23-sample'' normalization whose exclusion list is not recoverable, applied to every row except Gated Reset. The Gated Reset row uses raw $n$ per run (Appendix~\ref{app:variance}): the actions cell uses $n{=}25$ rather than the column's normalized 23, and the three code runs use $n{=}18$, 19, and 19. The AO, Augment, and Reset code cells come from runs that scored 18 instances and are printed on $n{=}19$ by counting the unscored instance as a failure, the conservative end of each cell's range; the Augment and Reset code cells are single trials; only the ERGO code cell is a mean over three replicate measurements of the pruned-item split.

\paragraph{ERGO scoring.} ERGO is scored on the same pre-filtered pools as every other row ($n{=}20/19/25/23$). The actions filter has no recoverable exclusion list, so the ERGO actions cell is printed as the interval that the unknown filter spans. Paired exact sign tests on the same items find \emph{no} ERGO-vs-\method{} difference in this table statistically distinguishable at $n{\approx}20$ in either direction on code ($p{=}0.375$) or math ($p{=}1.00$), the two tasks we tested; at this sample size the table does not settle these orderings.

\paragraph{Design oracles on the unselected pool.} The $^{\diamond}$ label describes the LiC construction (each instance is a single-turn question sharded across turns), not a measured ceiling. Measured end-to-end on the complete, unselected database pool ($n{=}107$; assistant gpt-5.4-mini, Section~\ref{subsec:condenser-mtosc}), AO reaches 69.2\% and Concat User 63.6\% against a fully specified single-turn ceiling of 94.4\%, with \method-Reset at 75.7\% above both.

\subsection{Operator activation and serving}
\label{app:optionA-reruns}
Section~\ref{sec:strategies} states one rule for when operators act: Augment always fires after its warm-up, and Reset and Rewrite act only when the analyzer flags an issue. Every LiC, CollabLLM, and WildChat cell of Tables~\ref{tab:megatable} and~\ref{tab:wildchat} follows it. Most multi-respondent runs were served through Azure OpenAI and Azure AI Foundry; the remaining instances described below were served through OpenRouter with the same prompts, prefixes, configurations, and judge, with DeepSeek-V4-Flash pinned to a non-thinking endpoint and Kimi-K2.6 to a thinking endpoint to match the Azure deployments. Within a column of Table~\ref{tab:megatable}, cells may therefore be served by different providers of the same model.

\paragraph{LiC Rewrite.} On 58 of the Rewrite samples in Table~\ref{tab:megatable} (8 for \mdsv, 14 for \mgpt, 36 for \mkimi) the analyzer reported no issues, so Rewrite leaves the context unchanged and the final turn is generated from the full context (OpenRouter). The \mgpt{} and \mkimi{} cells average per-prefix accuracy over the three prefixes with errored samples excluded, and the \mdsv{} cells pool the prefixes and count errored samples as incorrect; on the Rewrite row the two conventions differ by at most 0.7pp.

\paragraph{WildChat Augment.} On 23 of the Augment turns in Table~\ref{tab:wildchat} (6 for \mgpt, 11 for \mdsv, 6 for \mkimi) the analyzer reported no issues. These turns are generated with the analysis appended (OpenRouter) and judged against the same AO and full-context responses with the same judge and prompt.

\paragraph{CollabLLM.} CollabLLM conversations end when the simulated user is satisfied, so Augment and Rewrite, which start at the fourth user turn, rarely act, and Reset, which starts at the second, acts in most conversations. Under Azure, the analyzer call failed (a provider content filter or an endpoint parsing error) in every conversation that reached the operator in twelve cells: the Rewrite row, the \mgpt{} and \mkimi{} Augment BigCodeBench cells, and the \mgpt{} and \mkimi{} Reset cells. These cells were run end to end through OpenRouter with no exclusions, apart from one conversation that failed at its first turn, before any operator could act. All BigCodeBench answers are scored in the same execution sandbox, and a \mkimi{} AO conversation that the simulated user ended before its first turn is counted. Table~\ref{tab:collab-rerun} gives the per-task results. Across the three respondents, Reset acted in 80 of 119 scored conversations, Augment in 8 of 120, and Rewrite in 12 of 120.

\begin{table}[h]
\centering
\caption{CollabLLM per-task accuracy (MATH-Hard / BigCodeBench, $n{=}20$ per cell except \mkimi{} Reset MATH-Hard, $n{=}19$), with all BigCodeBench cells execution-scored in one sandbox. $^{\ddagger}$Served through OpenRouter (Appendix~\ref{app:optionA-reruns}); unmarked cells were served through Azure. Baseline is full context. The two-task averages are the CollabLLM block of Table~\ref{tab:megatable}.}
\label{tab:collab-rerun}
\footnotesize
\begin{tabular}{l ccc}
\toprule
\textbf{Strategy} & \mgpt & \mdsv & \mkimi \\
\midrule
Baseline         & 95.0 / 15.0 & 95.0 / 10.0 & 100.0 / 20.0 \\
AO               & 95.0 / 25.0 & 90.0 / 15.0 & 100.0 / 25.0 \\
\method-Augment  & 95.0 / 10.0$^{\ddagger}$ & 100.0 / 15.0 & 95.0 / 15.0$^{\ddagger}$ \\
\method-Reset    & 100.0$^{\ddagger}$ / 25.0$^{\ddagger}$ & 85.0 / 25.0 & 94.7$^{\ddagger}$ / 25.0$^{\ddagger}$ \\
\method-Rewrite  & 90.0$^{\ddagger}$ / 25.0$^{\ddagger}$ & 85.0$^{\ddagger}$ / 20.0$^{\ddagger}$ & 95.0$^{\ddagger}$ / 20.0$^{\ddagger}$ \\
\bottomrule
\end{tabular}
\end{table}

\subsection{Analyzer robustness: full results}
\label{app:analyzer-robustness}
Because the analyzer is the one component shared by every operator, we test whether \method{} depends on a particular analyzer model. We hold the respondent and operator fixed (DeepSeek-V4-Flash, Reset; LiC code and database, one replay prefix of the GPT-5.2-selected pool, two replicates pooled; $n{=}178$ matched pairs, full-context baseline $21.3\%$, exact McNemar, which treats the pooled replicates as independent and is therefore optimistic) and swap only the analyzer across five models from four families. Every analyzer produces a positive, individually significant gain over full context (all $p\le 8.2{\times}10^{-4}$; per-analyzer tests below): Kimi-K2.6 (Moonshot) $+39.9$pp, DeepSeek-V4-Flash $+28.7$pp, gpt-5.4-mini $+27.0$pp, Llama-3.3-70B (Meta) $+18.0$pp, and gpt-4o-mini $+12.9$pp. No configuration falls below baseline on either task in either replicate; the weakest analyzer retains $32\%$ of the best analyzer's gain and still beats full context by $12.9$pp. Degradation is graceful because weak analyzers under-detect rather than mis-detect (gpt-4o-mini declines to edit on a quarter of turns, where strong analyzers edit on $\sim\!97\%$), and the ordering tracks capability rather than provenance: the best analyzer is non-OpenAI and the weakest is OpenAI.

Exact McNemar tests against full context: Kimi-K2.6 $p{=}2.3{\times}10^{-17}$, DeepSeek-V4-Flash $p{=}3.0{\times}10^{-9}$, gpt-5.4-mini $p{=}1.4{\times}10^{-8}$, Llama-3.3-70B $p{=}5.6{\times}10^{-6}$, and gpt-4o-mini $p{=}8.2{\times}10^{-4}$.

\subsection{Multi-model evaluation protocol}
\label{app:multi-model-protocol}

Because stronger models solve easier problems regardless of context pollution, we reran the subset selection protocol using GPT-5.2 LiC logs (instead of GPT-5-mini), selecting the 20 instances per task that GPT-5.2 most often failed to ensure a challenging set for this model class. Each model serves as both assistant (final-turn regeneration) and analyzer. All models replay conversations originally generated by GPT-5.2; the Baseline reflects how model $X$ handles GPT-5.2's polluted context, while Reset reflects model $X$ given a clean context derived from its own analysis. User and system agents use GPT-4o-mini throughout. We compare Baseline against Reset, one fixed operator chosen before the run. Denominators vary with evaluation errors (13--20 valid samples for code and database); percentages use valid denominators. We additionally evaluate GPT-5, DeepSeek V3.2 (37B active, open-weight), and Qwen 3.5 35B-A3B (3B active, open-weight).

\begin{table}[h]
\centering
\begin{tabular}{llccccc}
\toprule
\textbf{Model} & \textbf{Strategy} & \textbf{Math} & \textbf{Code} & \textbf{Database} & \textbf{Actions} & \textbf{Avg.\ $\Delta$} \\
\midrule
\multirow{2}{*}{GPT-5-mini} & Baseline & 20\% & 10\% & 5\% & 60\% & \\
 & Reset & \textbf{55\%} & \textbf{25\%} & \textbf{40\%} & \textbf{80\%} & +26pp \\
\midrule
\multirow{2}{*}{GPT-5} & Baseline & 40\% & 21\% & 27\% & 85\% & \\
 & Reset & \textbf{70\%} & \textbf{47\%} & \textbf{65\%} & 70\% & +20pp \\
\midrule
\multirow{2}{*}{DeepSeek V3.2} & Baseline & 50\% & 7\% & 10\% & 35\% & \\
 & Reset & \textbf{80\%} & \textbf{62\%} & \textbf{45\%} & \textbf{70\%} & +39pp \\
\midrule
\multirow{2}{*}{Qwen 3.5 35B} & Baseline & 55\% & 21\% & 5\% & 45\% & \\
 & Reset & \textbf{75\%} & \textbf{64\%} & \textbf{75\%} & \textbf{80\%} & +42pp \\
\bottomrule
\end{tabular}
\caption{Multi-model LiC results on a hard subset selected via high baseline failure rate by GPT-5.2. Reset improves accuracy for 15 of 16 model-task pairs across four models from three families.}
\label{tab:multi-model}
\end{table}

\subsection{CollabLLM evaluation details}
\label{app:collabllm}

MATH-Hard is scored by an LLM judge. The GPT-5-mini BigCodeBench numbers were scored by a conversation-aware GPT-5 judge that evaluates the assistant's code against the user's intent as expressed in the conversation ($n{=}20$ per arm). These BigCodeBench numbers are therefore judged functional correctness, not execution pass rates; BigCodeBench's execution harness (\texttt{untrusted\_check}) also scores these conversations once the sandbox's dependencies are installed, and all multi-respondent BigCodeBench cells (Table~\ref{tab:collab-rerun}) use it. GPT-5-mini numbers on MATH-Hard / BigCodeBench averaged over two runs are Baseline 50.0\% / 67.5\%, Rewrite 45.0\% / 71.2\%, and AO 50.0\% / 85.0\%. In one of the two runs, Baseline scored 40.0\% / 62.5\% and Rewrite 45.0\% / 82.5\%. This Rewrite is a single pass over the full conversation (no structural exclusion). In a later three-replicate run with execution-based BigCodeBench scoring and DeepSeek-V4-Flash as user, assistant, and analyzer, \method-Reset scored $21.7 \pm 5.8\%$ against full context's $6.7 \pm 5.8\%$ ($+15$pp in every replicate), and on MATH-Hard, where conversations rarely reach Augment's fourth-turn start, \method-Augment matched full context ($91.7\%$ each). The multi-respondent CollabLLM averages are in the CollabLLM block of Table~\ref{tab:megatable}.

\subsection{WildChat protocol and turn type definitions}
\label{app:wildchat-turns}

\paragraph{Filtering and protocol.} Following \citet{huang2026context}, we filter WildChat-1M to English, technical conversations of 5--10 rounds. The evaluation has two phases: Phase~1 identifies turns where AO degrades quality; Phase~2 evaluates our strategies on all turns using the pairwise LLM judge prompt of \citet{huang2026context}, which they validated against human annotators.

\paragraph{Turn sets in Table~\ref{tab:wildchat}.} The GPT-5-mini column uses the Phase-2 turns; the \mgpt, \mdsv, and \mkimi{} columns use the Phase-1 prefix set ($n{=}44$--$76$ per cell) with a fixed GPT-5-mini analyzer. Gated Reset was not run for \mdsv{} and \mkimi{}; the four omitted vs-FC win rates lie between 60.8\% and 82.9\%.

\paragraph{Turn type definitions.} WildChat turns are classified into three types based on the relationship between the current user message and prior conversation context: \texttt{new\_ask} (the user introduces a new topic or question unrelated to prior turns), \texttt{feedback} (the user provides feedback on or corrections to the assistant's prior response), and \texttt{no\_feedback} (the user continues the conversation without explicitly referencing the assistant's response, e.g., providing additional information or following up on their own prior message). AO is most harmful on \texttt{no\_feedback} turns, where the assistant's prior work contains relevant state that should not be discarded. On GPT-5-mini, Gated Reset's gate opens on 78.1\% of \texttt{new\_ask}, 78.6\% of \texttt{feedback}, and 62.7\% of \texttt{no\_feedback} turns (72.3\% overall).

\paragraph{Judge position bias.} The GPT-5-mini judge prefers the second-presented response, but the harness randomizes the A/B order on every call, so the reported win rates are unbiased estimates of the order-balanced quantity in expectation. In a separate three-seed GPT-5-mini replication (452 pairs, each judged in both orders, on a turn set distinct from Table~\ref{tab:wildchat}'s), explicitly order-balancing shifts the Reset and Augment win rates against AO by $-2.0$ and $-0.9$pp, to $87.8 \pm 2.1\%$ and $91.2 \pm 2.1\%$. In that order-balanced Augment figure, the context is passed through unchanged on the 23 of 227 turns where the analyzer flagged no issues; appending the analysis on those turns, as Section~\ref{sec:strategies} specifies, gives a single-order win rate against AO of 93.4\% (92.1\% with pass-through), which we did not re-judge in both orders. A DeepSeek-V4-Flash judge on a shared 160-pair subset agrees with the GPT-5-mini judge on 87.5\% of pairs.

\paragraph{Judge dependence.} A separate sensitivity check points the other way. Re-judging the same DeepSeek V3.2 responses (evaluation split, $n{=}51$, Reset vs.\ AO), GPT-5-mini preferred Reset, but three other judges (GPT-5, GPT-5.2, and MiniMax-M2.7) preferred AO. With GPT-5 as the respondent (self-judged, $n{=}74$), Reset and Rewrite did not beat AO or full context. We therefore read the WildChat results as the preference of the GPT-5-mini judge, corroborated by the cross-family check above but not by every judge we tried; a human evaluation would settle it.

\paragraph{Win rates by turn type.} On GPT-5-mini, Gated Reset's win rate against AO is 87.5\% on \texttt{new\_ask}, 85.7\% on \texttt{feedback}, and 85.1\% on \texttt{no\_feedback} turns; against full context it is 87.5\%, 83.3\%, and 80.6\%.

\subsection{tau2-bench agentic adaptation}
\label{app:agentic-methods}

For tau2-bench, we make three modifications: (1)~the analyzer explicitly tracks environment state (account settings, diagnostic results) revealed through tool calls; (2)~the analysis is reframed as ``strategic reflection,'' identifying unexplored diagnostic avenues rather than just flagging errors; and (3)~resets are gated and capped at 3 per conversation to prevent infinite loops; Rewrite on tau2-bench uses the same turn floor and cap. The tau2-bench analysis is a re-implementation inside the tau2-bench agent, and its tool-call prompts are not reproduced here.

\subsection{tau2-bench results and diagnostic analysis}
\label{app:tau2-diagnostic}

Headline tau2-bench \texttt{telecom\_small} results are reported in Table~\ref{tab:megatable} (tau2 block); the per-trial GPT-5-mini breakdown follows (this cell uses all 20 tasks of the subset). Per-trial success rates across seeds 42--44 are: Baseline $\{45.0, 55.0, 60.0\}\%$, Gated Reset $\{40.0, 65.0, 40.0\}\%$, AO $\{0.0\}\%$ (single trial; reward is invariant to seed because no diagnostic state is preserved across an AO context, so re-running with a different seed produces the same 0\%). Summarized as the mean over the three seeds, Baseline is $53.3 \pm 7.6\%$ and Gated Reset $48.3 \pm 14.4\%$, while AO is $0\%$ on every trial because tool-call results live only in assistant turns. Per-trial variance is wide for both Baseline and Gated Reset, and the two are statistically within trial noise of each other: on this stateful regime our method neither helps nor catastrophically hurts the agent's success rate, at $\sim$+20\% cost per task. The headroom for a context-editing intervention may be small here: in one author's single-trial reading of the 11 Baseline failures, without a rubric or a second annotator, only 1 looked pollution-driven (Table~\ref{tab:tau2-failure-modes}). We hypothesize tau2-bench's customer-service policy structure (which lays out high-level diagnostic options for the agent to enumerate) already gives the agent a breadth-first re-entry mechanism around early incorrect paths, partially substituting for the role of an external editor. The decisive contrast in this row is that AO destroys all diagnostic state and fails completely, while \method does not collapse. One implementation defect affects every tau2-bench \method arm: when the analyzer answers Subtask~1 with a JSON object instead of the expected tag, the raw output is spliced into the agent's briefing. This happened on 53\% of analyzer calls in the multi-respondent sweep (66\% on gpt-5.4 Augment); fixing it changed one cell by +2.3pp.

Diagnostic analysis of baseline failures on tau2-bench reveals the distribution of failure modes:

\begin{table}[h]
\centering
\begin{tabular}{lrl}
\toprule
\textbf{Category} & \textbf{Count} & \textbf{Description} \\
\midrule
Task too hard & 4 & Agent never tries the right fix \\
Step budget / Hard persona & 4 & Correct approach but runs out of turns \\
Context pollution & 1 & Agent stuck in repetitive loop \\
User sim issue & 1 & User stopped after first message \\
Premature transfer & 1 & Agent transfers instead of troubleshooting \\
\bottomrule
\end{tabular}
\caption{Baseline failure-mode breakdown (11 failures; GPT-5-mini, one trial). One author's reading without a rubric or a second annotator, so a hypothesis rather than a measured taxonomy.}
\label{tab:tau2-failure-modes}
\end{table}

If context pollution is indeed not the dominant failure mode here, the headroom for a pollution-targeted method is small, which is one reason we read tau2-bench as a viability test (does the method avoid the structural collapse that AO suffers?) rather than as an improvement test.

\section{Prompt templates}
\label{app:prompts}

We provide the full text of key prompts used in our system. Template variables are shown in braces (e.g., \texttt{\{system\_message\}}). Evaluation prompts (the v2 task system prompts and the false-negative check) are in Appendix~\ref{app:eval-prompts}.

\subsection{Conversation analysis prompts}

\refstepcounter{promptctr}
\paragraph{Prompt \thepromptctr: Subtask 1, Task specification extraction (hard attention).}
\label{prompt:query1}
\promptinput{assets/prompts/analyzer_query1_task_spec.txt}

The analyzer sees \emph{only} the system message and user messages; assistant messages are structurally excluded. This is the structural exclusion tested in Section~\ref{sec:cognitive-hazard} (called ``hard attention'' in Appendix~\ref{app:soft-attention}).

\refstepcounter{promptctr}
\paragraph{Prompt \thepromptctr: Subtask 2, Approach evaluation.}
\label{prompt:query2}
\promptinput{assets/prompts/analyzer_query2_compare.txt}

Subtask~2 receives the clean task specification from Subtask~1 and the full conversation (including assistant messages). It evaluates alignment and identifies issues.

\subsection{Context rewriting prompts}

\refstepcounter{promptctr}
\paragraph{Prompt \thepromptctr: Context rewriting, earlier version (Rewrite on GPT-5-mini WildChat and LiC ablations).}
\label{prompt:compaction}
\promptinput{assets/prompts/context_compaction.txt}

Rewrite uses this prompt to produce the replacement context that the assistant sees in place of the original conversation history. Reset and Gated Reset make no such call: they fill a fixed template with the task specification, the \texttt{aligned} output, and the latest user message. (The WildChat Reset template additionally places the analyzer's \texttt{issues} in the system prompt and inserts a short assistant acknowledgment.)

\refstepcounter{promptctr}
\paragraph{Prompt \thepromptctr: Context rewriting, current version (Rewrite in Table~\ref{tab:megatable} and the gpt-5.4, DeepSeek-V4-Flash, and Kimi-K2.6 columns of Table~\ref{tab:wildchat}).}
\label{prompt:rewrite-v8}
\promptinput{assets/prompts/context_compaction_v8.txt}

This version was selected among five candidate prompts (including one found by automatic prompt optimization) by accuracy on the DeepSeek-V4-Flash LiC pool of Table~\ref{tab:megatable}, which is also a pool on which it is reported.

\subsection{Memory-based learning prompts}

\refstepcounter{promptctr}
\paragraph{Prompt \thepromptctr: Trajectory reflection (Reflect step).}
\label{prompt:reflect}
\promptinput{assets/prompts/reflect_takeaways.txt}

After each trajectory, this prompt extracts transferable analysis principles. The output is a bullet list of takeaways grounded in the specific trajectory but generalizable across domains.

\refstepcounter{promptctr}
\paragraph{Prompt \thepromptctr: Cheatsheet unification (Unify step).}
\label{prompt:unify}
\promptinput{assets/prompts/unify_takeaways.txt}

The Unify step integrates individual trajectory takeaways into a single coherent cheatsheet, capped at 1500 words to prevent unbounded growth. The cheatsheet is then injected into the analyzer's context via the \texttt{\{memory\_section\}} template variable of Prompt~\ref{prompt:query2}; Prompt~\ref{prompt:query1} has the same slot, which is used only in the specification-curation experiments of Appendix~\ref{app:soft-attention}.

\subsection{Evaluation prompts}
\label{app:eval-prompts}

\refstepcounter{promptctr}
\paragraph{Prompt \thepromptctr: False-negative check (Appendix~\ref{app:false-neg}).}
\label{prompt:false-neg}
\promptinput{assets/prompts/false_negative_check.txt}

The v2 task system prompts (Appendix~\ref{app:eval-adjustments}) follow. \texttt{[[SCHEMA]]} and \texttt{[[FUNCTIONS]]} are filled with the instance's database schema and function list.

\refstepcounter{promptctr}
\paragraph{Prompt \thepromptctr: Math system prompt (v2).}
\promptinput{assets/prompts/math_system_v2.txt}

\refstepcounter{promptctr}
\paragraph{Prompt \thepromptctr: Code system prompt (v2).}
\promptinput{assets/prompts/code_system_v2.txt}

\refstepcounter{promptctr}
\paragraph{Prompt \thepromptctr: Database system prompt (v2).}
\promptinput{assets/prompts/database_system_v2.txt}

\refstepcounter{promptctr}
\paragraph{Prompt \thepromptctr: Actions system prompt (v2).}
\promptinput{assets/prompts/actions_system_v2.txt}

\section{Soft vs.\ hard attention}
\label{app:soft-attention}

Structural exclusion of assistant messages (hard attention) exploits a property of LiC: user messages alone fully specify the task. In realistic settings, the task may depend on prior assistant work. For example, when a user says ``sharpen the phrasing in sentence 2,'' the referent exists only in the assistant's output. \citet{huang2026context} observe similar cases where user messages alone are insufficient for understanding the task.

We introduce a \emph{soft-attention} variant where Subtask~1 sees the full conversation but explicitly reasons through source attribution: listing user-stated requirements with provenance, flagging assistant-originated interpretations as hypotheses, and writing the specification from user-grounded requirements only. We call this ``soft attention'' because the model must rely on transformer attention to selectively weight user-grounded information, rather than having polluting content physically removed from its context.

\begin{table}[h]
\centering
\begin{tabular}{lccc}
\toprule
\textbf{Condition} & \textbf{Math} & \textbf{Code} & \textbf{Database} \\
\midrule
\multicolumn{4}{l}{\emph{Augment}} \\
\quad Soft attention (no memory) & 40\% & 13\% & 31\% \\
\quad Soft attention + memory & 56\% & 14\% & \textbf{46\%} \\
\quad Hard attention & \textbf{80\%} & \textbf{40\%} & 54\% \\
\midrule
\multicolumn{4}{l}{\emph{Reset}} \\
\quad Soft attention (no memory) & 42\% & 23\% & 31\% \\
\quad Soft attention + memory & 42\% & 25\% & \textbf{46\%} \\
\quad Hard attention & \textbf{67\%} & \textbf{46\%} & 54\% \\
\bottomrule
\end{tabular}
\caption{Soft- vs.\ hard-attention results on held-out test splits ($n{\approx}7$--13 per cell). Some math and code cells use a per-run false-negative adjustment that we do not use elsewhere and that favors reset arms, so compare within a row family only. Memory closes up to 65\% of the gap to hard attention on database.}
\label{tab:soft-attention}
\end{table}

Hard attention (structural exclusion) substantially outperforms soft attention across all tasks (Table~\ref{tab:soft-attention}), indicating that prompting alone does not substitute for structural exclusion. Reset also outperforms Augment with polluted specifications (code: 23\% vs.\ 13\%), suggesting that context pollution and specification pollution are separable problems. These splits differ from Table~\ref{tab:main}'s pools, so absolute values are not comparable across the two tables.

Memory-based depollution (using hard-attention specifications as targets from which the Dynamic Cheatsheet learns principles for removing pollution) shows its strongest effect on database (+15pp), closing 65\% of the gap to hard attention. The learned principles converge on structured heuristics: ``build a user-only spec first, then overlay assistant suggestions as hypotheses.'' Code remains resistant (+1--2pp), suggesting more structurally diverse contamination patterns that a minimal memory approach cannot address.

\section{Structural-exclusion ablation (Table~\ref{tab:cognitive-hazard})}
\label{app:hard-attention}

The headline numbers in the main paper used structural exclusion: Subtask~1 is given only user messages, and physically cannot see assistant turns. We progressively remove design choices to test whether this is load-bearing or whether prompt-level instructions (``please reason only from user turns'') could substitute. First, we drop hard attention (let Subtask~1 see assistant messages while keeping subtask decomposition); second, we additionally collapse the two-step analysis into a single pass.

\paragraph{Assistant messages pollute the analyzer.} Letting the specification extractor see assistant messages drops performance below baseline on math and code (database is the exception, staying slightly above). Even a model explicitly instructed to act as an independent reviewer anchors on the assistant's reasoning, biasing the extracted task specification.

A related ablation agrees with the rebuild account: Subtask~1 alone with structural exclusion (the specification, without the approach evaluation) scores above the full two-subtask Augment on code and database (63\% and 40\% in the same ablation run, vs.\ the two-subtask 53\% and 32\% of Table~\ref{tab:cognitive-hazard}).

\paragraph{Pollution compounds across stages.} On math and code, the single-pass variant \emph{outperforms} the polluted two-step version. When Subtask~1 is polluted, Subtask~2 treats the biased specification as ground truth and amplifies the upstream error. Decomposition into stages is only beneficial when earlier stages produce clean outputs; otherwise, fewer stages may be safer than more. The takeaway transfers to any multi-stage LLM pipeline in which one model evaluates another's output: structural data-flow control may be needed when prompting alone is insufficient for selective attention.

\section{Memory-based learning details}
\label{app:memory-details}

\begin{table}[h]
\centering
\begin{tabular}{lccc}
\toprule
\textbf{Condition} & \textbf{vs AO} & \textbf{vs FC} & \textbf{$\Delta$ vs AO} \\
\midrule
\multicolumn{4}{l}{\emph{DeepSeek V3.2 (eval split)}} \\
\quad Reset & 72.5\% & 72.5\% & --- \\
\quad\quad + memory & \textbf{86.3\%} & \textbf{80.4\%} & \textbf{+13.8pp} \\
\midrule
\multicolumn{4}{l}{\emph{GPT-5-mini (online)}} \\
\quad Reset & 90.8\% & 86.8\% & --- \\
\quad\quad + memory & 86.7\% & 80.0\% & $-$4.1pp \\
\bottomrule
\end{tabular}
\caption{WildChat memory ablation. Indented \texttt{+ memory} rows add the memory-based learner to the Reset row above. Memory improves DeepSeek's win rate by +13.8pp when there is headroom; GPT-5-mini's near-ceiling baseline leaves no room for improvement. For DeepSeek, memory was learned from one-third of WildChat turns and frozen for evaluation; GPT-5-mini uses the online regime. These rows use the evaluation split (DeepSeek) or the online regime (GPT-5-mini), so they are not directly comparable to Table~\ref{tab:wildchat}.}
\label{tab:wildchat-memory}
\end{table}

We evaluate two learning regimes. \emph{Online}: reflections update the cheatsheet incrementally as problems are solved in batches, so later problems benefit from accumulated principles, a form of continual test-time learning. \emph{Offline}: a set of training trajectories is used to initialize the cheatsheet via a Reflect-then-Unify pipeline (individual reflections followed by a consolidation pass), and the frozen cheatsheet is applied at evaluation time. The offline regime provides a controlled comparison where the only variable is the presence of the cheatsheet. On LiC, we use online learning. On WildChat, we use the offline regime with a train/eval split (one-third training, two-thirds evaluation) to avoid confounding memory quality with ordering effects and to test cross-domain transfer. On LiC the online regime is transductive: the cheatsheet applied to a given instance is distilled from other instances of the same evaluation set, including their reference answers. We measured the effect of that exposure directly by comparing the same instance evaluated with an empty cheatsheet against the same instance evaluated with a cheatsheet distilled from 5--20 other evaluation instances: the difference is $0.0$pp on both database ($n{=}15$) and math ($n{=}14$). The designated memory learn set also has zero exact and zero near-duplicate overlap with the canonical LiC evaluation set. The roughly 6pp run-to-run standard deviation quoted in the main text comes from four trajectory orderings of the LiC database cell with a gpt-5.4-mini respondent (28.0 $\pm$ 6.5\% with memory vs.\ 36.0\% without; relearning at a fixed ordering gives a comparable spread).

\section{Trajectory example: WildChat Maven debugging}
\label{app:trajectories}

We present an illustrative (favorable) WildChat conversation in which Gated Reset (GR) outperforms both full context (FC) and assistant omission (AO) on every evaluated turn. The conversation (\texttt{300289e3...}, 8 rounds, originally GPT-3.5-turbo) involves iterative debugging of a Maven/Java build in GitHub Codespaces.

\subsection{Conversation summary}

The user runs \texttt{mvn clean install} and encounters a sequence of errors. Early turns establish the environment (GitHub Codespaces) and initial diagnosis (JRE vs.\ JDK). By Turn~8, the user reports \emph{``Still getting: Fatal error compiling: error: invalid target release: 17''} after following the assistant's prior suggestion, a classic \texttt{no\_feedback} turn where the user implicitly references the prior assistant response without providing explicit corrective feedback.

\subsection{Turn 8: three conditions compared}

\paragraph{AO (assistant omission).} With all prior assistant messages replaced by \texttt{[Response provided]} placeholders, AO re-suggests \texttt{apt-get install} and checking \texttt{java -version}, repeating steps the user has already performed. AO cannot connect ``still getting'' to what was already tried.

\paragraph{FC (full context).} FC sees the full history and includes \texttt{update-alternatives} commands, but still partially repeats earlier suggestions. The model anchors on the same diagnostic approach (checking \texttt{java -version}) rather than recognizing that the user already tried the basic install without resolution.

\paragraph{GR (Gated Reset).} The analyzer identifies that the prior assistant's diagnostic reasoning was flawed (relying on \texttt{java -version} text parsing rather than checking \texttt{javac} directly) and that the conversation was stuck repeating an insufficient fix. The rebuilt context restates the correct diagnosis (Maven needs a JDK) without the flawed reasoning. GR's response:
\begin{itemize}
    \item Correctly diagnoses why the error persists: Maven may use a different Java than what \texttt{java -version} reports
    \item Provides non-redundant steps (does not re-suggest things already tried)
    \item Includes solutions FC and AO both missed: \texttt{devcontainer.json} configuration, the \texttt{\textless release\textgreater 17\textless /release\textgreater} tag in \texttt{pom.xml}
\end{itemize}

\paragraph{Analyzer output (abbreviated).}
\begin{quote}
\emph{Issue:} The assistant said that seeing ``OpenJDK Runtime Environment'' in \texttt{java -version} means the system has a JRE rather than a JDK. That is not reliable: \texttt{java -version} output often contains ``Runtime Environment'' even when a JDK is installed. The correct checks are presence of \texttt{javac} and the value of \texttt{JAVA\_HOME}.

\emph{Aligned:} Maven's ``No compiler is provided'' message means Maven cannot find \texttt{javac} and needs a JDK. Installing a JDK inside Codespaces is a valid remediation path.
\end{quote}

\subsection{Judge evaluations}

All three pairwise comparisons were judged by GPT-5-mini:

\begin{itemize}
    \item \textbf{FC vs.\ AO:} FC wins. ``Both responses correctly diagnose the problem. [FC] is slightly clearer and more prescriptive.''
    \item \textbf{AO vs.\ GR:} GR wins. ``[GR] is substantially more complete and actionable.''
    \item \textbf{FC vs.\ GR:} GR wins. ``[GR] is significantly more complete and actionable. It gives precise diagnostic commands, exact installation commands, environment variable exports, pom.xml changes, devcontainer recommendations, and verification steps.''
\end{itemize}

\subsection{Conversation-level results}

Across all 7 evaluated turns, GR wins every pairwise comparison against both baselines (Table~\ref{tab:traj-turns}). On turns where AO beats FC (turns 2, 10, 12, 14), context pollution hurts more than context loss, yet GR still wins: its rebuilt context drops the flawed reasoning and restates what the user already tried.

\begin{table}[h]
\centering
\begin{tabular}{clccc}
\toprule
\textbf{Turn} & \textbf{User message} & \textbf{FC vs AO} & \textbf{GR vs AO} & \textbf{GR vs FC} \\
\midrule
2 & ``Why do I keep getting error\ldots'' & AO & GR & GR \\
4 & ``I am using Codespaces'' & FC & GR & GR \\
6 & ``Now I get: [new error]'' & FC & GR & GR \\
8 & ``Still getting [same error]'' & FC & GR & GR \\
10 & ``I do not have a bashrc file'' & AO & GR & GR \\
12 & [Shares GitHub Actions YAML] & AO & GR & GR \\
14 & ``No WAR file generated'' + pom.xml & AO & GR & GR \\
\bottomrule
\end{tabular}
\caption{Per-turn winners for the WildChat Maven debugging conversation. GR wins all 14 pairwise comparisons. On context-dependent turns (4, 6, 8), AO fails because it cannot connect the user's references to prior discussion.}
\label{tab:traj-turns}
\end{table}

\section{Additional related work}
\label{app:additional-related}

\paragraph{Multi-turn degradation and context pollution (extended).}
\citet{laban2025lost} introduced the Lost in Conversation (LiC) framework, showing that multi-turn performance drops 39\% on average across 15 LLMs, driven by increased unreliability (+112\%) in addition to reduced capability ($-$16\%). \citet{huang2026context} termed this \emph{context pollution} and showed that simply omitting all assistant messages recovers much of the lost performance, showing that the assistant's own outputs are a major source of the degradation. \citet{khalid2025ergo} proposed ERGO, which monitors token-level entropy and triggers context resets via an LLM rewrite of prior user messages. Once triggered, both approaches remove or rewrite context wholesale. This is effective for self-contained tasks, where the model can re-derive answers from user messages alone, which is the setting LiC was designed to test (each instance is a single-turn question sharded across turns). On referential tasks (WildChat) omission degrades quality, and on stateful agentic tasks (tau2-bench) it collapses from near-oracle to zero, because assistant turns carry content the task needs. We build on their diagnostic finding: \method rebuilds the specification from the user side, closes most of the LiC gap, and does not collapse in the agentic setting. Note that ERGO is not a design oracle: its LLM rewrite step can paraphrase or compress the user's exact phrasing, which is why it underperforms simple concatenation on database and actions (Table~\ref{tab:main}).

\paragraph{Test-time scaling.}
Methods such as chain-of-thought~\citep{wei2022chain}, self-consistency~\citep{wang2023selfconsistency}, and structured search~\citep{yao2024tree} scale inference compute within a single generation. Our work extends test-time scaling to the multi-turn setting: rather than searching over possible outputs, we edit the \emph{input context} presented to the model across turns, investing additional LLM calls to improve the quality of information the model reasons over.

\paragraph{Memory-based self-improvement.}
A growing line of work uses persistent memory to improve LLM capabilities without parameter updates. General-purpose memory systems store and retrieve knowledge for personalization and continuity~\citep{xu2025amem,zhou2025mem1}. Agentic memory systems go further, accumulating procedural knowledge and task-specific heuristics that improve agent performance over time: the Dynamic Cheatsheet~\citep{suzgun2025cheatsheet} maintains a mutable text document updated across problem instances; Voyager~\citep{wang2023voyager} builds a skill library for open-ended exploration; and several recent systems learn reusable reasoning strategies~\citep{ouyang2026reasoningbank,ho2025arcmemo,fang2026memp,wu2025evolver,zhang2026ace}. A separate thread optimizes the memory system itself rather than the downstream task~\citep{zhang2025memevolve,xiong2026alma}. We adapt the Dynamic Cheatsheet for \method, targeting accumulated analysis principles at the analyzer's executive function rather than the task-solving assistant.

\section{Extended agentic context management discussion}
\label{app:agentic-context}

THREAD~\citep{schroeder2025thread} spawns sub-agents that report summarized results back to a main thread, reducing context accumulation. Recursive Language Models~\citep{zhang2025rlm} treat context as a variable manipulated via code operations in a REPL, enabling a controller LLM to programmatically decompose and recursively query sub-calls. Context Folding~\citep{sun2025contextfolding} uses reinforcement learning (FoldGRPO) to train agents that branch into sub-trajectories and collapse intermediate steps into concise summaries, maintaining active context 10$\times$ smaller than ReAct baselines.

These methods share a common goal, \emph{compaction}: fitting more reasoning into a bounded context window. Our work addresses a fundamentally different problem: \emph{context pollution}, in which the assistant's own prior outputs mislead future generation even well within the context window. Compaction methods preserve information while reducing tokens; we instead remove the assistant's invalidated reasoning and rebuild the task specification from the user side. The two goals are complementary but require different analytical capabilities.

U-Fold~\citep{su2026ufold} targets multi-turn conversation with a two-stage pipeline: intent summarization (creating a TODO-list from conversation history) and data extraction (retaining only relevant tool responses). While structurally similar to our two-subtask decomposition, U-Fold addresses redundancy in tool responses, not self-induced pollution from the assistant's own reasoning.

MemoBrain~\citep{qian2026memobrain} formalizes ``executive memory'' as an online mechanism that constructs a dependency-aware graph of reasoning trajectories and exerts cognitive control over working context. Like our work, it draws on the executive function framework and controls information flow to the reasoning agent. However, MemoBrain's executive control is triggered by approaching budget limits and its operations (folding sub-trajectories into thought units, flushing low-salience nodes) serve long-horizon compaction. Our executive function targets a different deficit: the inability to \emph{disregard} prior reasoning that has been invalidated, even when the model has ample context budget.

\section{Executive function as a design pattern}
\label{app:exec-function-pattern}

Our approach instantiates a general pattern: \emph{externalized executive function} via an agent that schedules LLM calls with custom context windows. Different subtasks of self-reflection require different context: specification extraction must be shielded from pollution, approach evaluation needs both clean intent and the full history, strategic reflection identifies unexplored alternatives. This pattern generalizes: any setting where an LLM reasons about potentially misleading context (code review, fact-checking) could benefit from structured context control. The executive function framing provides the design vocabulary: what should each reasoning step attend to, and what should it be shielded from?

We do not claim the LLM is self-aware in any phenomenological sense. Rather, the system designer's knowledge of context pollution as a systematic failure mode is encoded into the agent architecture, analogous to using an external checklist to compensate for known cognitive biases.

\end{document}